\documentclass{article}

\usepackage[final]{corl_2018}

\usepackage{times}
\usepackage{epsfig}
\usepackage{graphicx}
\usepackage{float}
\usepackage{wrapfig}

\usepackage{amsmath,amssymb}

\usepackage{bm,xspace}
\usepackage{comment}
\usepackage{verbatim}
\usepackage{multirow}
\usepackage{balance}
\usepackage{url}
\usepackage{booktabs}
\usepackage{etoolbox,siunitx}
\usepackage{calc}
\usepackage{pifont,hologo}
\usepackage{color}

\newcommand{\ra}[1]{\renewcommand{\arraystretch}{#1}}

\usepackage[super]{nth}
\usepackage{nicefrac}
\def\aa{\mathbf{a}}

\def\oo{\mathbf{o}}

\def\eps{\varepsilon}

\DeclareMathSymbol{@}{\mathord}{letters}{"3B}

\definecolor{alexey}{rgb}{0.8,0.,0.8}

\def\latex/{\LaTeX}
\def\bibtex/{\hologo{BibTeX}}

\usepackage{graphicx}
\usepackage{booktabs}
\usepackage{verbatim}
\usepackage{float,array}

\graphicspath{{./images/}}

\title{Motion Perception in \\ Reinforcement Learning with Dynamic Objects}

\author{
  Artemij Amiranashvili\\
  University of Freiburg
  \And
  \hspace{-5mm}
  Alexey Dosovitskiy\\
  \hspace{-5mm}
  Intel Labs
  \AND
  \hspace{7mm}
  Vladlen Koltun\\
  \hspace{7mm}
  Intel Labs
  \And
  Thomas Brox\\
  University of Freiburg
}

\begin{document}
\maketitle


\begin{abstract}
In dynamic environments, learned controllers are supposed to take motion into account when selecting the action to be taken. However, in existing reinforcement learning works motion is rarely treated explicitly; it is rather assumed that the controller learns the necessary motion representation from temporal stacks of frames implicitly. In this paper, we show that for continuous control tasks learning an explicit representation of motion improves the quality of the learned controller in dynamic scenarios. We demonstrate this on common benchmark tasks (Walker, Swimmer, Hopper), on target reaching and ball catching tasks with simulated robotic arms, and on a dynamic single ball juggling task. 
Moreover, we find that when equipped with an appropriate network architecture, the agent can, on some tasks, learn motion features also with pure reinforcement learning, without additional supervision. 
Further we find that using an image difference between the current and the previous frame as an additional input leads to better results than a temporal stack of frames.\footnote{This is an extended version of the \href{http://proceedings.mlr.press/v87/amiranashvili18a.html}{CoRL paper} (2nd Conference on Robot Learning (CoRL 2018), Z\"urich, Switzerland)
with the additional image difference baseline~\cite{wang2016temporal}. }
\end{abstract}



\section{Introduction}
\label{sec:introduction}

In many robotic tasks, the robot must interact with a dynamic environment, where not only the dynamics of the robot itself but also the unknown dynamics of the environment must be taken into account. 
Examples of such tasks include autonomous driving, indoor navigation among other mobile agents, and manipulation of moving objects such as grasping and catching.
The presence of moving elements in the environment typically increases the difficulty of a control task substantially, necessitating fast reaction time and prediction of the future trajectories of the moving objects.

In deep reinforcement learning (DRL), using a neural network as function approximator, a model of the environment's dynamics can, in principle, be learned implicitly. 
In simple cases, such as in some Atari games, corresponding motion features seem to be picked up automatically~\citep{Mnih2015}. 
However, it can be observed that a model operating on just a single frame often has the same performance as a model that takes a stack of successive images as input~\cite{DosovitskiyKoltun2017}. 
Is motion uninformative or is it just harder to learn than static features for an end-to-end trained system? 
Intuitively, we expect the latter, but then: how can we best enable the use of motion when training controllers? 

In this paper, we confirm the importance of motion in learning tasks that involve dynamic objects, and we investigate the use of optical flow to help the controller learn the use of motion features. 
In a straightforward manner, optical flow can be just provided as an additional input to an RL agent. 
A complication with this approach is that accurate optical flow computation is typically too slow for training of RL models, which requires frame rates of at least hundreds of frames per second to run efficiently. 
To address this issue, we design a small specialized optical flow network derived from FlowNet~\citep{Dosovitskiy2015}. 
The network is small enough to be run jointly with reinforcement learning while keeping computational requirements practical. 
We consider two training modes: one where the optical flow network is trained in a supervised manner beforehand, and one where the same network is trained online via RL based just on the rewards, i.e., without explicit supervision on the optical flow.

We perform extensive experiments on multiple diverse continuous control tasks.
We observe that the use of optical flow consistently improves the quality of the learned policy. 
The improvement is higher the more relevance the dynamics have for the completion of the task. 
Some tasks involving dynamic objects cannot be learned at all without the explicit use of motion. 
In some tasks unsupervised learning of the optical flow based on the rewards is possible, whereas on harder tasks, direct supervision is still required to kickstart the motion representation learning.

We also find that a network provided with the current frame concatenated with the difference between the current and the previous frame outperforms or matches the image stack baseline across all tasks with dynamic objects. This suggests that the image difference could be a more useful input for pixel control reinforcement learning than the usually used stack of frames~\citep{Mnih2015}. 

\section{Related Work}
\label{sec:related_work}

Deep reinforcement learning aims to learn sensorimotor control directly from raw high-level sensory input via direct maximization of the task performance, by using deep networks as function approximators.
This approach has allowed learning complex behaviors based on raw sensory data in various domains: arcade game playing~\citep{Mnih2015}, navigation in simulated indoor environments~\citep{Mnih2016,Mirowski2017,DosovitskiyKoltun2017,Jaderberg2017}, simulated racing~\citep{Mnih2016}, simulated robotic locomotion~\citep{Lillicrap2016} and manipulation~\citep{Andrychowicz2017,Tobin2017}, as well as manipulation on physical systems~\citep{Gu2017}.
Despite these notable successes, there is little understanding of how and what exactly do the DRL agents learn.
In this work, we focus on studying how DRL makes use of motion information in dynamic environments.

Previous works in DRL vary in how they provide motion information to the network.
The most standard approach is to feed a stack of several recent frames to the agent, assuming that the deep network will extract the motion information from these if needed~\citep{Mnih2015,Mnih2016,DosovitskiyKoltun2017}.
On the architecture side, agents are commonly equipped with a long short-term memory (LSTM) that can, in principle, pick up the motion information~\citep{Mnih2016,Mirowski2017,Jaderberg2017}.
An alternative approach to using motion information is based on future frame prediction, which can be used to learn a useful feature representation~\citep{Finn2016} or to plan future actions explicitly~\citep{FinnLevine2017,Ebert2017}. 
A representation of motion and static properties can also be learned through a self-supervised multi-frame TNC embedding~\citep{dwibedi2018learning}.
In contrast to all these works, we aim to understand what representation of motion is the most useful for an RL agent and in particular experiment with explicitly computed optical flow.

The use of optical flow relates our work to the line of research on using explicit perception systems to improve the performance of learned sensorimotor control policies. 
Providing ground truth depth maps to the agent has been shown to lead to improved navigation performance compared to a system making use of only color images~\citep{Mirowski2017,Savva2017}.
In the domain of autonomous driving, semantic segmentation can help improve the driving command prediction~\citep{Xu2017} or allows the transfer from simulation to the real world~\citep{Mueller2018}.
\citet{Goel2018} show that object segmentation learned in an unsupervised fashion leads to improved performance in some Atari games.
\citet{Clavera2017} use object detection to improve transfer of learned object manipulation policies.
Our work is similar in spirit to these, but we focus on analyzing the use of motion and optical flow in deep reinforcement learning, which, to our knowledge, has not been previously addressed.

While optical flow is not commonly used in DRL, it has a long history in robotics.
Vision-based robotic systems have employed optical flow a range of diverse applications: tracking~\citep{Luo1988}, navigation~\citep{Muratet2004,VardyMoller2005,Chao2014}, obstacle avoidance~\citep{SouhilaKarim2007}, visual servoing~\citep{Allen1991}, object catching~\citep{SuShaojie2016}.
Applications of optical flow have been complicated by the trade-off between computational efficiency and the accuracy.
Only recently, deep-learning-based methods have allowed for fast and accurate estimation of optical flow~\citep{Dosovitskiy2015}.
In this paper, we build on this progress and use a miniaturized variant of FlowNet~\citep{Dosovitskiy2015,IMSKDB17} to estimate optical flow.
Our optimized small FlowNet is extremely efficient, which allows its use for training reinforcement learning agents.

\section{Method}
\label{sec:method}

We study an agent operating in an environment in discrete time.
At each time step $t$ the agent gets an observation $\oo_t$ from the environment and generates an action $\aa_t$ in response.
In this work we focus on environments where observation is a high-dimensional sensory input, such as an image, and the action is a relatively low-dimensional vector of continuous values.
In addition to the observation, at each step the agent gets a scalar reward $r_t$.
In this work the reward is often the sum of two terms $r_t = r_t^{sc} + r_t^{sh}$: the typically sparse scoring reward $r_t^{sc}$ (we often refer to it as score) and a denser shaping reward $r_t^{sh}$.
We are interested in achieving high scoring reward, but add a shaping reward to simplify training.

Since we deal with continuous control tasks, we use Proximal Policy Optimization (PPO)~\citep{schulman2017proximal} as our base RL algorithm.
To enable processing of high-dimensional inputs, we use a convolutional network (CNN) as a function approximator.
We use an architecture similar to~\citet{Mnih2015}.
In tasks involving manipulation of moving objects we provide the vector of robot state variables to the network in addition to the high-dimensional sensory observation.
We process this vectorial input by a separate fully connected network and concatenate the output with the output of the perception part of the CNN (full architecture is shown in Table \ref{tbl:convnet}).

\begin{figure}[t]
	\centering
    \includegraphics[width=0.95\linewidth]{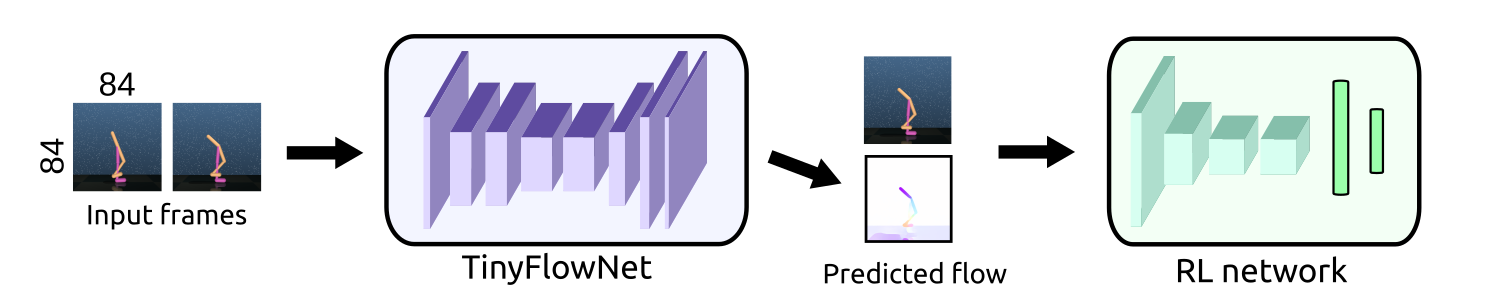}
	\caption{Illustration of the approach. The RL agent uses an explicit motion representation provided in the form of optical flow.}
	\label{fig:architecture}
	\vspace{-2mm}
\end{figure}

To understand the role of motion perception in training of an RL agent, we vary the input provided to the agent.
The straightforward options are to provide the network with just the current observation or several recent observations stacked together.
A more interesting scenario is to provide optical flow explicitly to the RL network.
In this case, we use a separate convolutional network to estimate the optical flow.
This setup is illustrated in Figure~\ref{fig:architecture}. 

For optical flow estimation we use a miniaturized version of the FlowNetS network~\cite{Dosovitskiy2015}, which we refer to as TinyFlowNet.
This is necessitated by two considerations: first, we need the flow computation to be sufficiently fast to support RL training and, second, the input resolution used for RL is much smaller than that assumed by the full FlowNet.
TinyFlowNet consists of a 5-layer encoder and a 2-layer decoder, compared to a 9-layer encoder and a 4-layer decoder in the original FlowNet.
Moreover, there are only two strided layers, the maximum number of channels is $128$, and all convolutional kernels are $3\times3$. 
We find that this smaller network is sufficiently expressive to accurately estimate optical flow in environments we consider in this work, while processing $1800$ image pairs per second on a Geforce 1080 Ti GPU. The full TinyFlowNet architecture is shown in Table \ref{tbl:tinyflownet}.

\begin{figure}[t]
	\centering
	\vspace{-2mm}
	\includegraphics[width=0.78\linewidth]{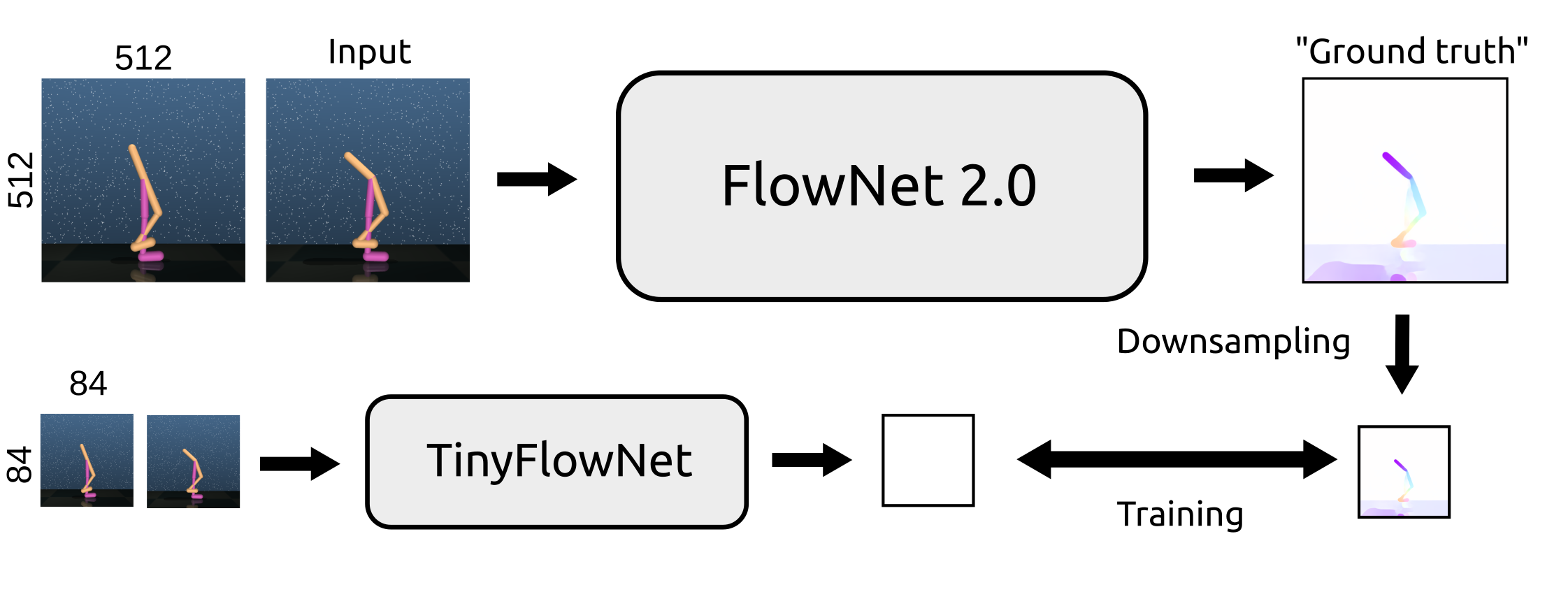}\\
	\vspace{-3mm}
	\caption{Training of TinyFlowNet using FlowNet 2.0~\citep{IMSKDB17} as a teacher.}
	\label{fig:flow_pretraining}
	\vspace{-2mm}
\end{figure}

We investigate two approaches to training the two-network system: pre-training the flow network separately or training both networks from scratch with RL.
In the first case, we pre-train TinyFlowNet in a supervised fashion on data extracted automatically from the RL environments using FlowNet 2.0~\citep{IMSKDB17} to provide targets for training; see Figure~\ref{fig:flow_pretraining}.
This student-teacher setup allows training without ground truth optical flow, making the approach  applicable to arbitrary environments.
In the second case, we initialize both networks with random weights and train the whole system from scratch with RL.

\subsection{Training details}
We use images of resolution $84 \times 84$ pixels as sensory observations in all environments.
The action space varies depending on the environment.
We train all agents for $20$ million time steps.
This is longer than what is typically used for PPO~\cite{schulman2017proximal}, since training from raw sensory observations is more difficult than from low-dimensional state vectors. We use the same hyperparameters as used by~\citet{schulman2017proximal} for Atari environments. However, we adjust the learning rate to $1 \times 10^{-4}$ and the number of epochs to $2$, which resulted in better and more stable performance in our environments.

The pre-training of TinyFlowNet is illustrated in Figure~\ref{fig:flow_pretraining}. 
We compute optical flow in high resolution ($512\times 512$ pixels) using FlowNet 2.0.
This optical flow is downsampled to $84\times 84$ pixels and used as target for training TinyFlowNet. 
To ensure accurate optical flow prediction, we trained a separate flow network for each of the environments, by extracting a dataset of $20@000$ image pairs.
For the standard control tasks (Walker, Swimmer, Hopper) we execute random actions to generate training data.
In our new tasks with moving objects, we keep the robot arm static while creating the dataset. 
This makes the optical flow estimation focus on the moving objects.

\section{Experiments}
\label{sec:experiments}

We compare the flow-based approach against several baselines on standard control tasks and on a series of new tasks that require interaction with dynamic objects. 
We evaluated the following models:
\begin{itemize}
    \item \textbf{Image:} processes the current image by a feedforward CNN
    \item \textbf{Image stack:} processes a stack of the $2$ most recent images by a feedforward CNN
    \item \textbf{Image difference:} processes the current image stacked with the image difference between the previous image and the current image
    \item \textbf{LSTM:} processes the current image by a CNN with an LSTM layer
    \item \textbf{Segmentation:} processes the current image and a segmentation mask of the moving object by a feedforward CNN. The mask is a motion segmentation taken from the predicted optical flow
    \item \textbf{Flow:} processes the current image and the optical flow between the current frame and the previous one by a feedforward CNN. Flow is computed in the backward direction to ensure that the object in the flow image is co-located with the object in the color image
    
\end{itemize}

\subsection{Standard control tasks}
\begin{figure}[t]
	\centering
	\setlength{\tabcolsep}{1mm}
	\begin{tabular}{ccc}
    \includegraphics[width=0.32\linewidth]{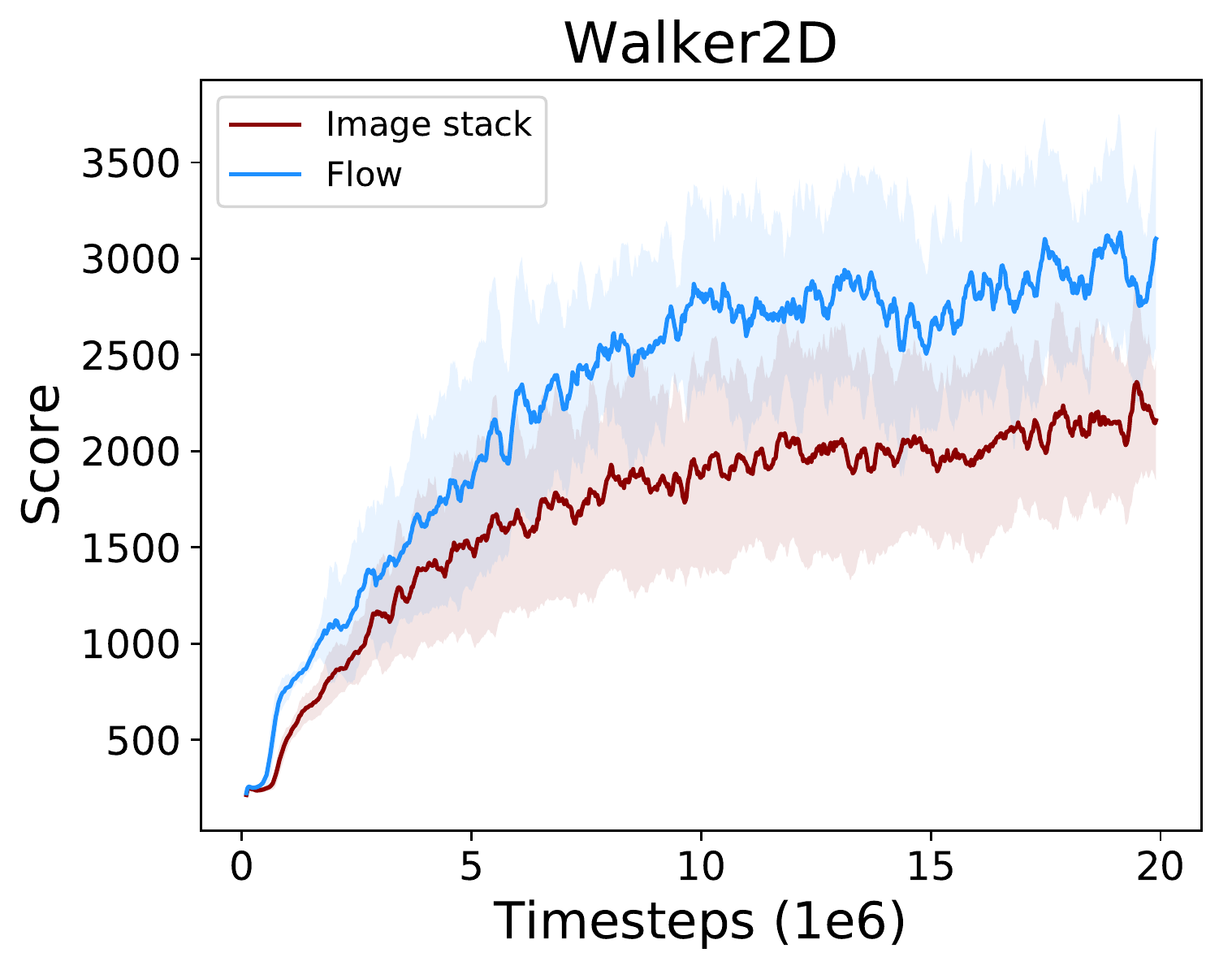} &
    \includegraphics[width=0.32\linewidth]{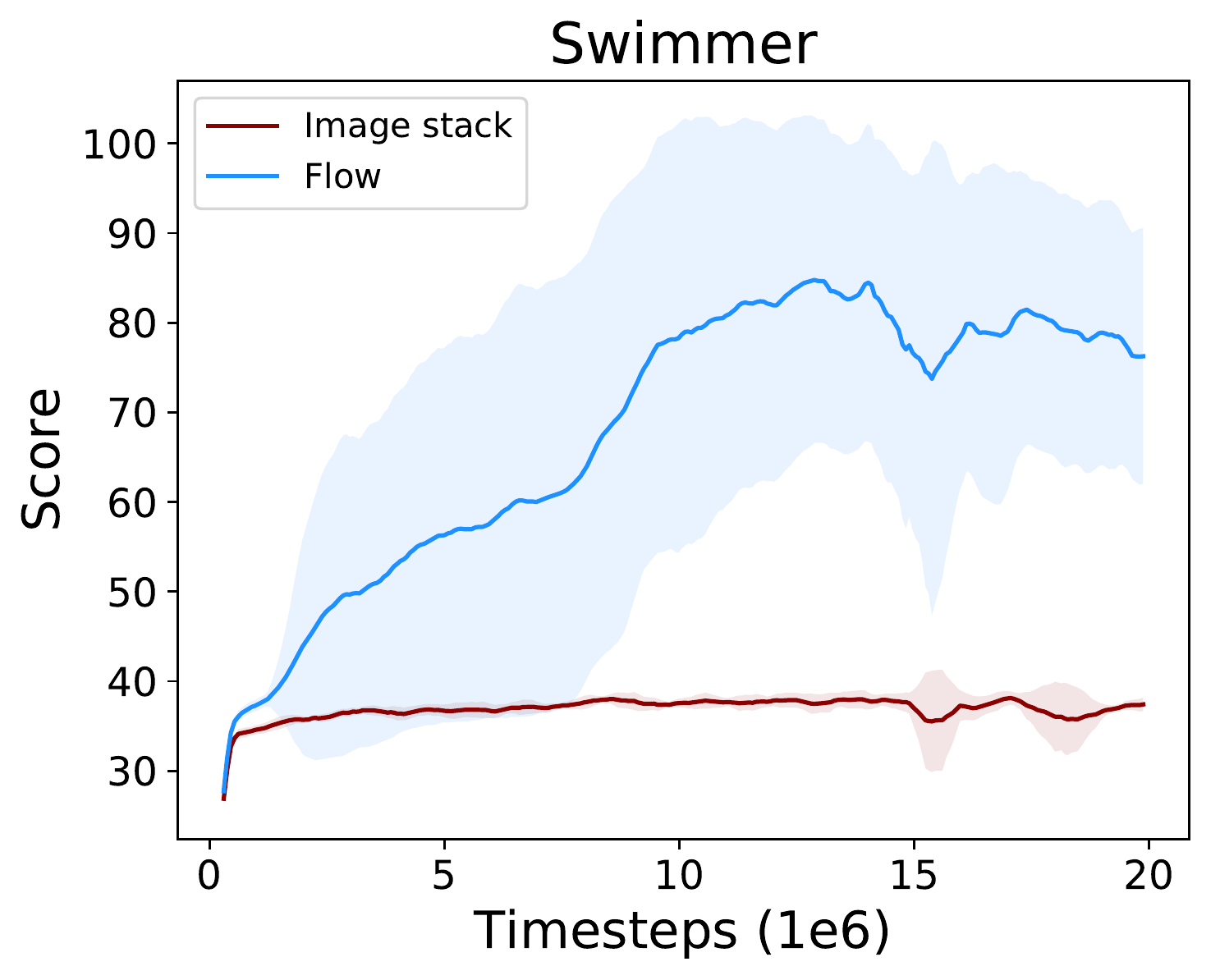} &
    \includegraphics[width=0.32\linewidth]{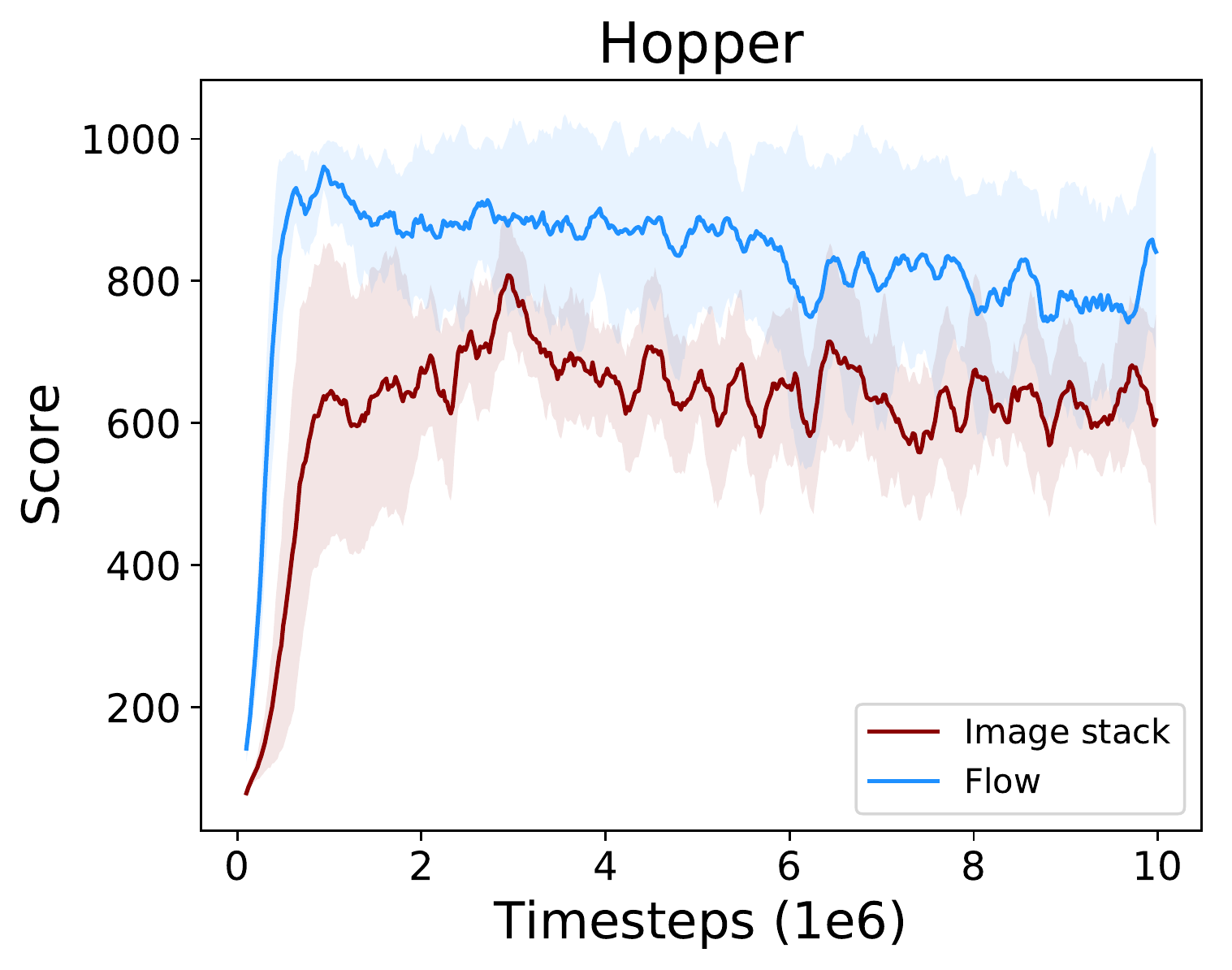}\\
    \end{tabular}
	\caption{Training curves on standard control tasks with pixel control. We trained 8 models in each condition. Lines show the mean reward; shaded areas show the standard deviation.}
	\label{fig:standard_control_curves}
\end{figure}

We started by experimenting with three standard control tasks from the OpenAI Gym framework~\citep{brockman2016openai}: Walker, Swimmer, and Hopper. 
We additionally adjusted these environments with visual modifications from the DeepMind Control Suite~\citep{tassa2018deepmind}.
Typically, these tasks are trained with the robot's state vector provided as input to the network.
We rather focused on learning solely from raw images and investigated whether information about motion, represented by optical flow, helps learning better policies. 
Because of the high variance of the performance on these tasks~\cite{henderson2017deep}, we trained each model 8 times with different random seeds and show the average performance and the standard deviation in Figure~\ref{fig:standard_control_curves}.

Although there are no moving objects in these tasks apart from the agent itself, providing optical flow as input clearly improves results compared to providing just the stack of images. This supports our initial hypotheses that motion information is very useful in dynamic environments and that the agent has problems deriving good motion features from the plain image stack using a standard network architecture.

\subsection{Tasks with dynamic objects}

We analyze the effect of motion perception in more detail on a specifically designed set of tasks, where the environment surrounding the robot contains moving objects.
In such environments, the use of motion information is expected to be even more crucial than on the control tasks above. 
In these experiments we complement the high-dimensional sensory observations with the vector containing the current state of the robot.

\begin{figure}[b]
	\centering
	\setlength{\tabcolsep}{1mm}
	\begin{tabular}{cccc}
	\includegraphics[height=0.23\linewidth]{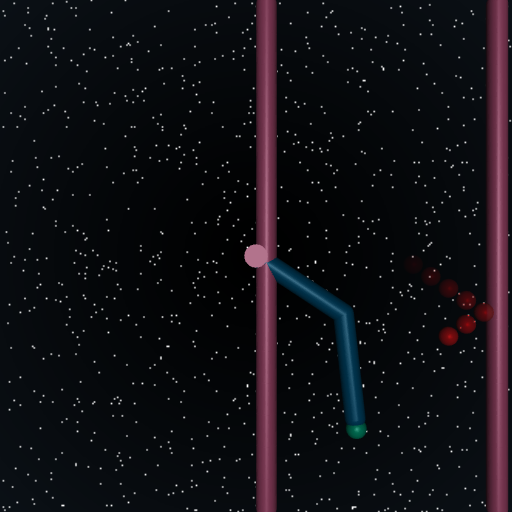} &
    \includegraphics[height=0.23\linewidth]{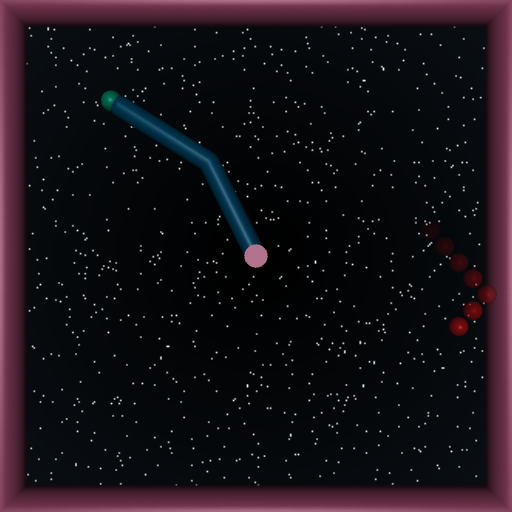} &
    \includegraphics[height=0.23\linewidth]{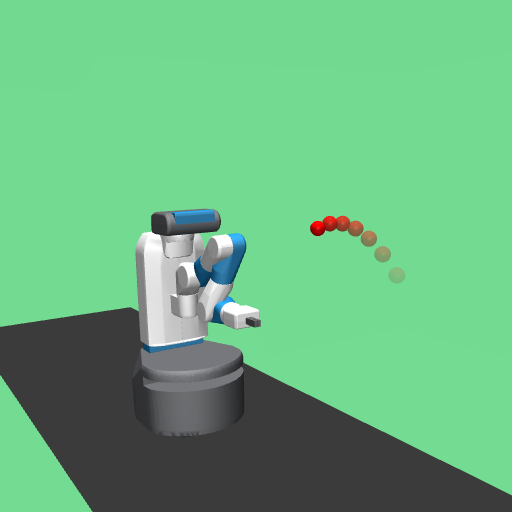} &
    \includegraphics[height=0.23\linewidth]{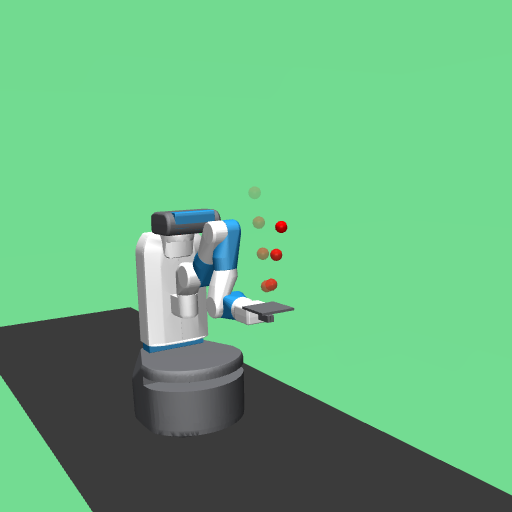} \\
    (a) 2D Catcher & (b) 2D Chaser & (c) 3D Catcher & (d) 3D KeepUp \\ 
    \end{tabular}
	\caption{The tested environments with moving objects.}
	\label{fig:dynamic_environment_tasks}
\end{figure}

We implemented four such environments in the MuJoCo simulator~\cite{todorov2012mujoco} by modifying OpenAI Gym tasks~\citep{brockman2016openai}.
Two of these are set up in a two-dimensional space and two in a three-dimensional space.
The environments are illustrated in Figure~\ref{fig:dynamic_environment_tasks}. All tasks terminate after 250 time steps.
\begin{itemize}
    \item \textbf{2D Catcher.} A 2-link 2D robotic arm is fixed in the center of the field as in the standard reacher environment. The target is a ball moving from the top of the screen towards the bottom, reflecting from two walls like in billiard. The aim is to ``catch'' the target by making the end effector of the arm overlap with the target. After the ball is caught a new target appears from the top.
    \item \textbf{2D Chaser.} A 2-link 2D robotic arm is fixed in the center of the field as in the previous task, but the target now reflects from the four borders. The aim is to keep the end effector of the robotic arm as close to the target as possible while the target keeps moving.
    \item \textbf{3D Catcher.} A 3-link 3D robotic arm is fixed on a base. Moving targets follow randomized parabolic trajectories in the vicinity of the arm. The aim is to ``catch'' the target with the end effector of the arm.
    \item \textbf{3D KeepUp.} A 3-link 3D robotic arm is fixed on a base and has a square pad fixed on its end effector. A ball falls down from the top under the effect of gravity. The aim is to reflect the ball with the pad and keep reflecting it every time it falls, by moving the arm and rotating the pad.
\end{itemize}

Like in the standard MuJoCo control tasks, each reward function also contains a motion penalty term to reduce unnecessary movement of the robot arm.
Further details are provided in the sections below. 
Environment configuration files, reward parameters, implementations, and a video showing the tasks and qualitative results will be made available on the project page:~{\small\url{https://lmb.informatik.uni-freiburg.de/projects/flowrl/}}.

\begin{figure}[t]
	\centering
	\setlength{\tabcolsep}{1mm}
	\begin{tabular}{rr}
	\includegraphics[height=0.38\linewidth]{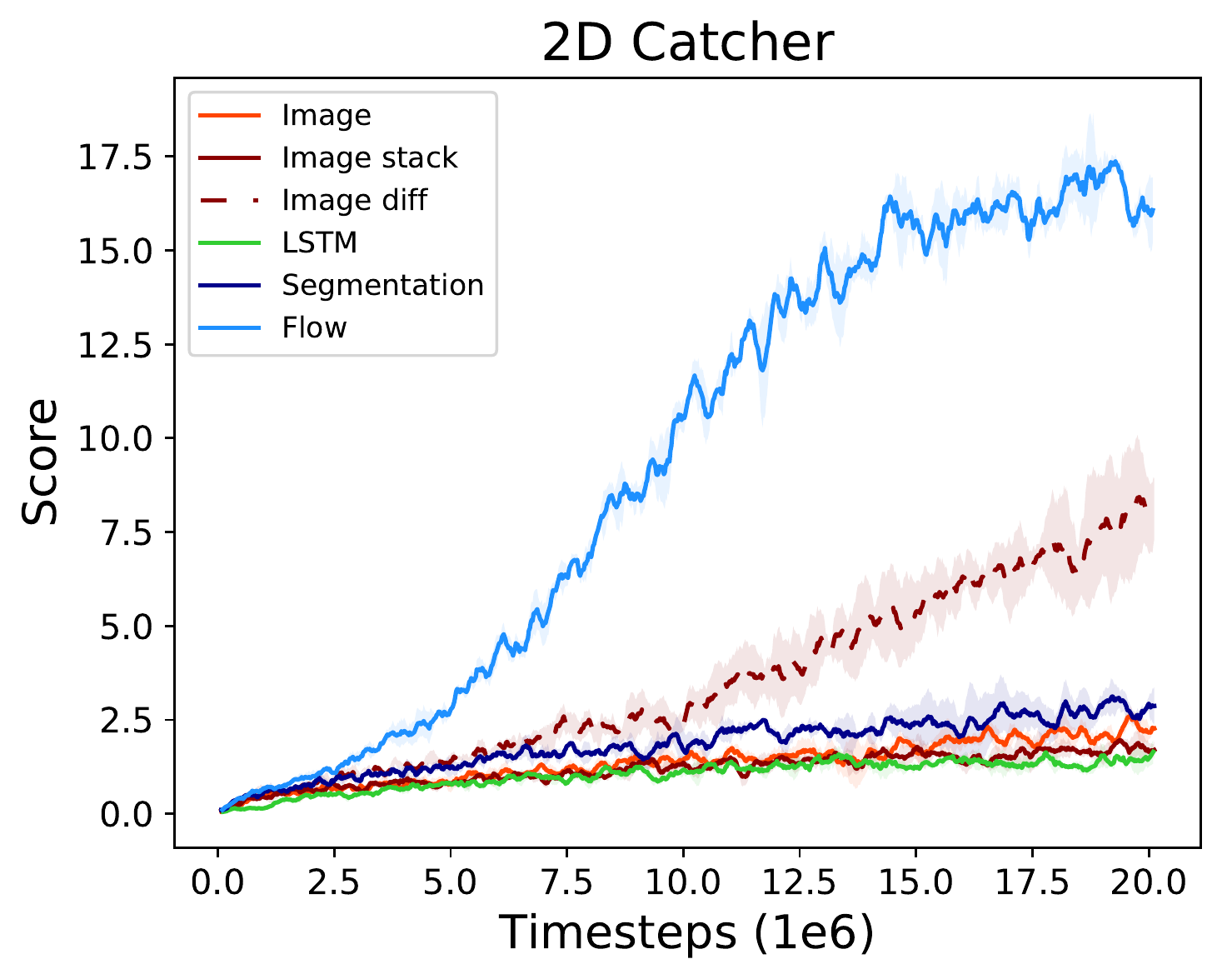} &
    \includegraphics[height=0.38\linewidth]{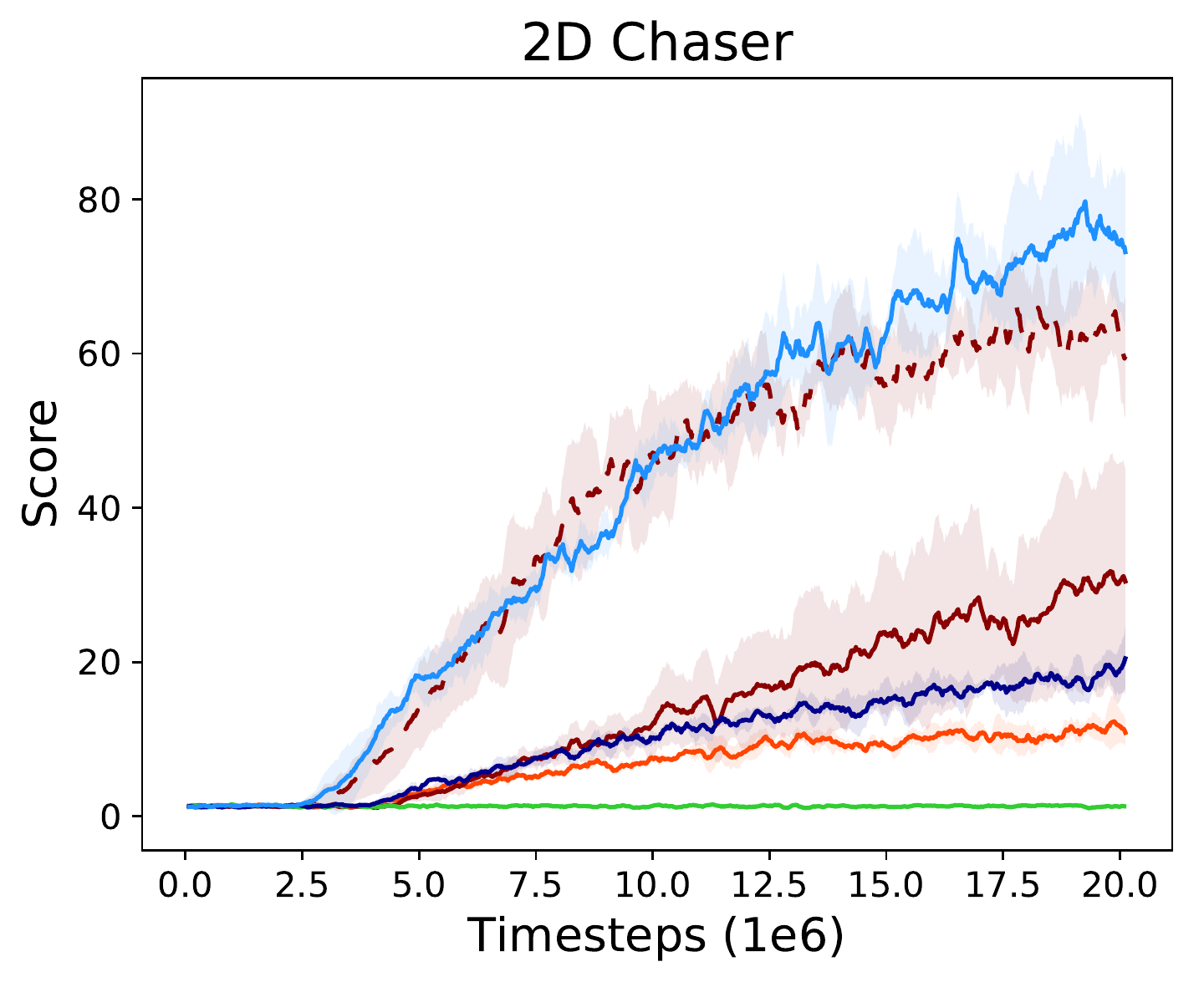} \\
    \includegraphics[height=0.38\linewidth]{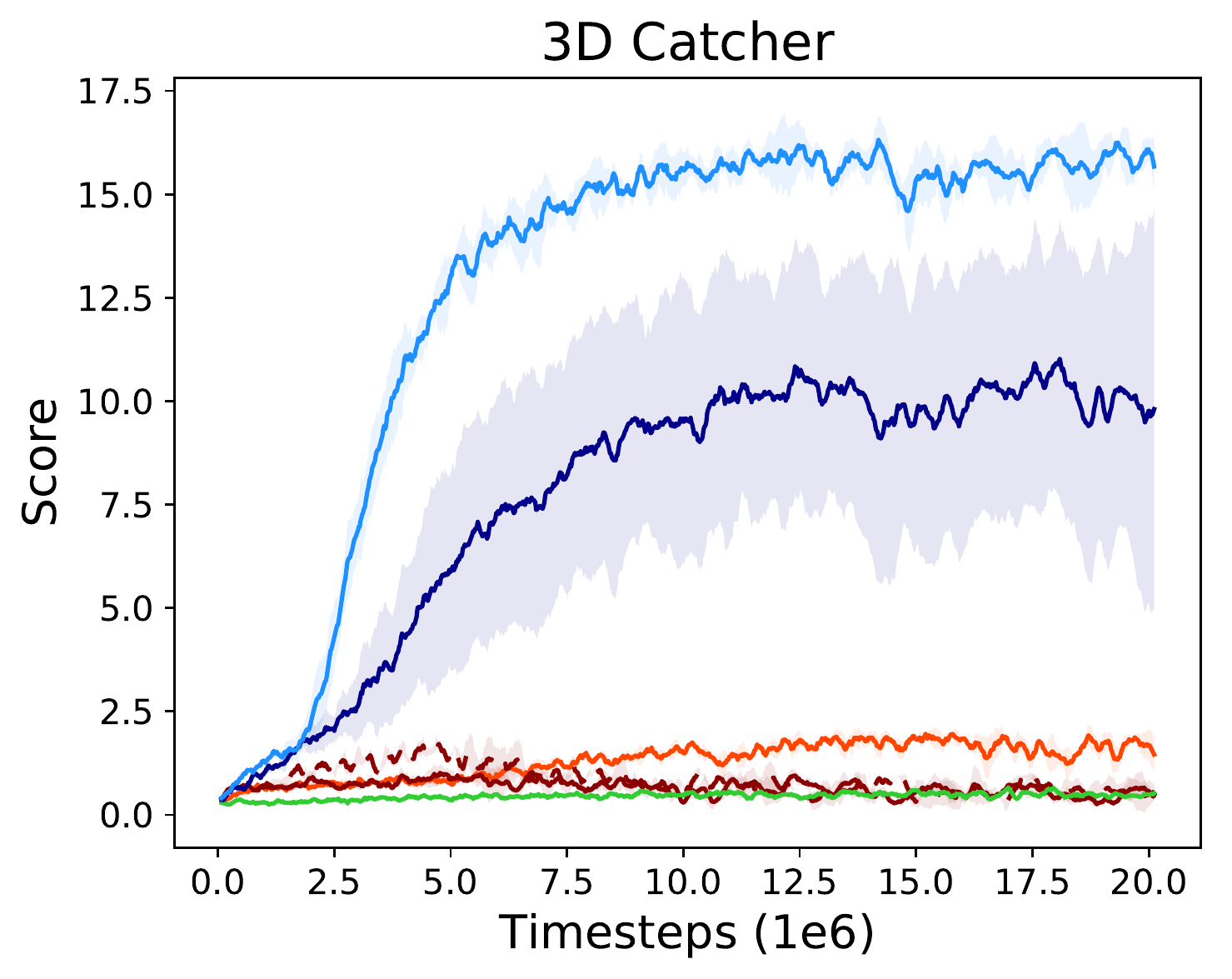} &
    \includegraphics[height=0.38\linewidth]{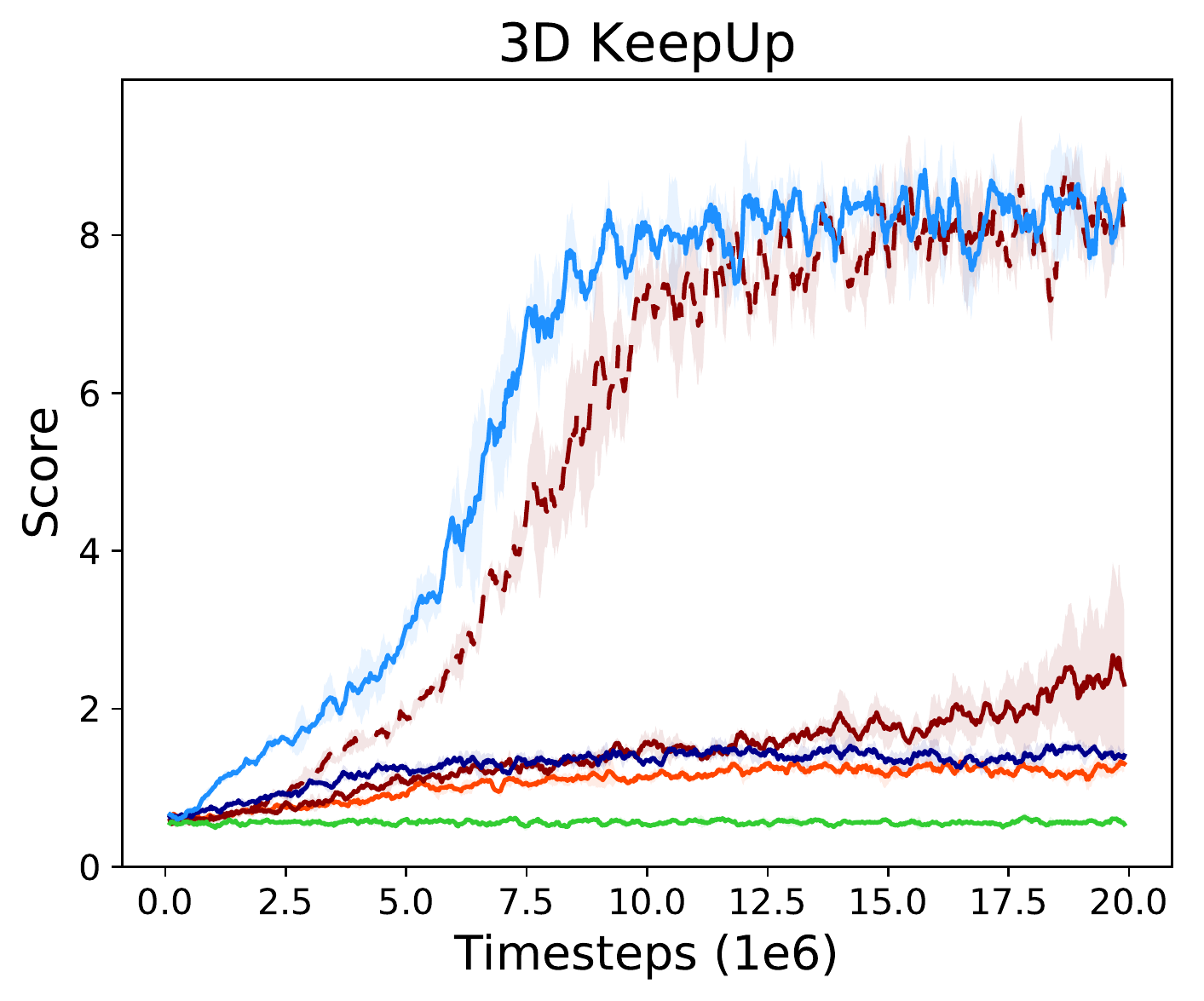} \\
    \end{tabular}
	\caption{Performance on the four tasks that involve moving objects. Overall the agent that uses optical flow outperforms other baselines, including LSTM and an agent provided with a stack of two recent frames.}
	\label{fig:dynamic_environment_results}
\end{figure}

\paragraph{2D environments.}
In 2D environments the robots are controlled by applying torque at the joints. 
In both tasks the agent receives a dense shaping reward depending on the distance to the target. In addition, it receives a sparse scoring reward when the distance between the end effector and the target falls below a fixed threshold (corresponding to overlap of the end effector with the target). In case of the \emph{2D Catcher}, this counts as a catch and a new ball is spawn, while in case of \emph{2D Chaser} the ball keeps moving.

The achieved scores are shown in Figure~\ref{fig:dynamic_environment_results} (top).
On both tasks, the use of optical flow improved the performance of the agent. Most alternative strategies that would allow the agent to use motion information, such as LSTM units or the image stack, hardly improved over the use of a single image. 
However, the image difference baseline clearly outperformed other baselines and nearly matched the performance of the flow-based agent on the \emph{2D Chaser} task.
It is interesting that a minor change in the input representation from image stack to the image difference leads to such a dramatic performance improvement, despite the fact that a network with image stack as input could easily learn to compute the image difference.
We believe this inability to learn the simple difference operation is due to the complexity and unstability of the network optimization.

Providing the segmentation mask of the moving ball did not reach the same performance as providing the optical flow. This shows that the optical flow is not just used for localizing the moving object, but also for predicting its future position. This is particularly important for the \emph{2D Catcher} task, where the agent easily misses the ball without a good prediction of the future ball position. The arm is not fast enough to catch up with the falling ball when it was missed.

\paragraph{Varying the target speed.}
\begin{wrapfigure}{r}{5.8cm}
    \centering
    \includegraphics[width=5.8cm]{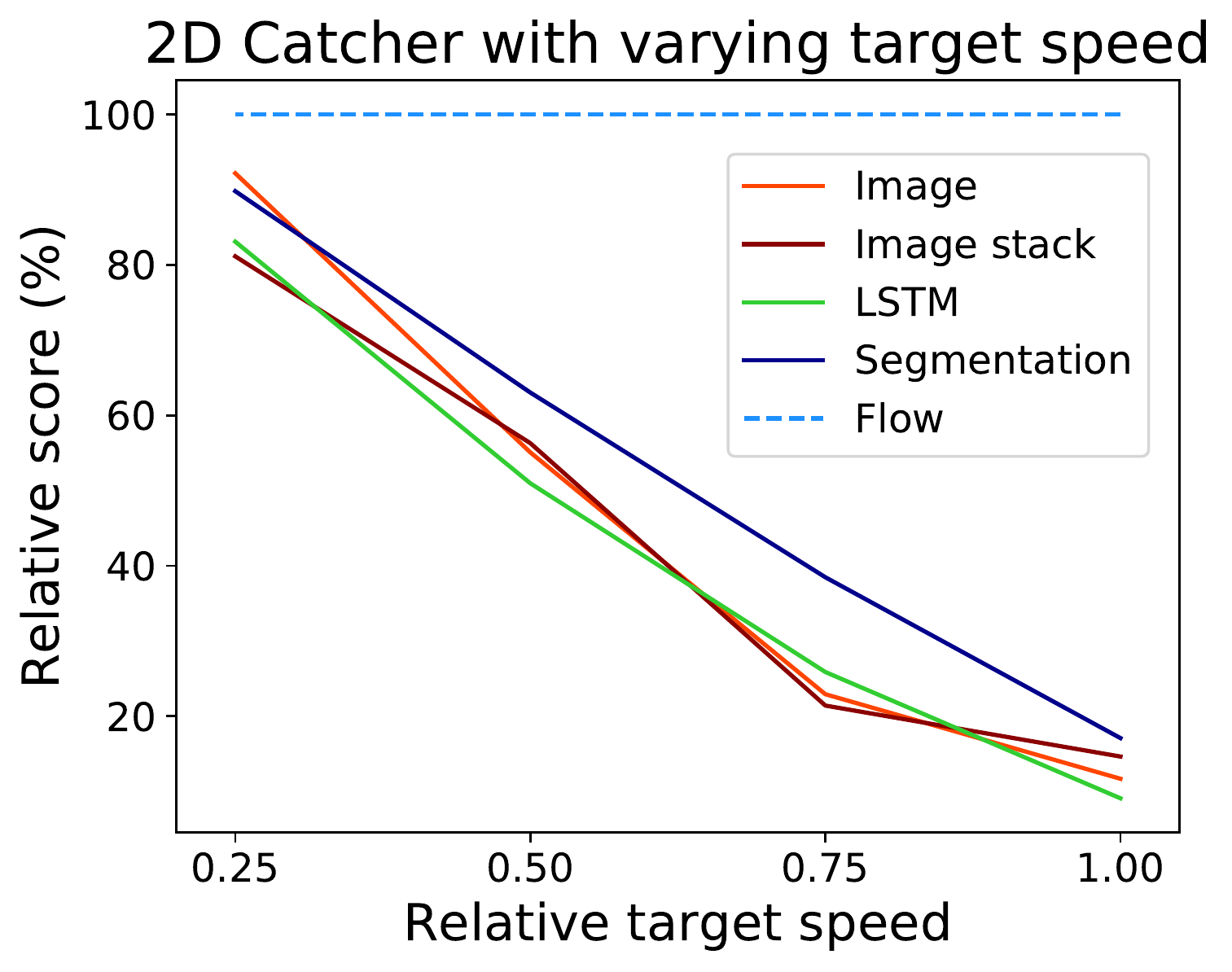}
    \vspace{-5mm}
    \caption{Results on the \emph{2D Catcher} task when varying the speed of the target. We plot the score relative to an agent equipped with optical flow.
    }
    \vspace{-5mm}
    \label{fig:catcher_varying_speed}
\end{wrapfigure}
The faster the motion in the environment relative to the robot's speed, the more important is the ability to plan ahead and, in order to do so, to estimate the motion of the objects. We performed an experiment to verify this hypothesis empirically.
We varied the speed of the target in the \emph{2D Catcher} task and measured the scores.

Figure~\ref{fig:catcher_varying_speed} shows the relative performance of the baselines to the flow-based agent as a function of the speed of the target.
As expected, the slower the target, the closer the performance of all methods.
However, even for slow targets the flow-based agent has a small advantage.

This might be because even for slow targets motion information helps catching them slightly faster, or, alternatively, because optical flow is not only useful for predicting the future trajectory of objects, but also for detecting moving objects which is useful even if the objects are slow. 

\paragraph{3D environments.}
In the 3D environments we provide two perpendicular camera views to the agent for it to have sufficient perceptual information to act in 3D space (shown in Figure \ref{fig:robot_view}). 
The agent must combine the information from both views to control the end effector relative to the target in 3D space. 
The robots in these environments are position controlled. 
In the case of the \emph{3D Catcher} the action space is 3-dimensional and includes the movement along the $x$, $y$, and $z$ axis. The shaping reward is the distance to the target future location in the plane of the end effector. The agent scores for each catch.   
In the \emph{3D KeepUp} task the shaping reward is the distance along the x-y plane between the target and the middle of the square pad. The agent scores each time it successfully reflects the target.

The results are shown in Figure~\ref{fig:dynamic_environment_results} (bottom).
In both cases, the flow-based agent learned effective policies, while the agent provided with an image, an image stack, or a LSTM layer could not solve the task. The agent with a motion segmentation mask outperformed other baselines on the \emph{3D Catcher} task, but could not reach the score of the flow-based agent.
The image difference baseline matched the performance of the flow-based agent on the \emph{3D KeepUp} task.

\paragraph{Analysis of motion representations.}
In order to better understand the effect of motion representations on learning, we experiment with providing the agent with a low-dimensional velocity vector of the target instead of per-pixel optical flow. We compute the velocity vector from the optical flow prediction and feed it to the RL agent as an additional vector input. We also measure the performance of the RL agent with ground truth optical flow or velocity vector. The results are shown in Figure \ref{fig:flow_vs_motion_vector_results}. Overall, the agent with access to per-pixel optical flow outperforms the velocity vector input. The agent with optical flow ground truth performs better in the 2D environments, indicating that the TinyFlowNet results could potentially be improved by using a larger network with better optical flow prediction.

\subsection{Learning motion features with deep RL}

The previous experiments show that availability of a pre-trained explicit optical flow estimator  improves the agent's performance on dynamic tasks. The typical network architecture used in most RL works, even when equipped with LSTM units, is not able to learn a good motion representation just from the reward signals. Is this still true if we train RL from scratch with a more powerful network?

We experiment with two larger network architectures. 
The first one is the one used in experiments with pre-trained optical flow: a TinyFlowNet with a normal RL network on top, but trained end-to-end from scratch.
The second one is a residual network~\cite{he2016deep} with 8 convolutional layers and approximately the same number of parameters as the combination of the TinyFlowNet with the normal RL network (the exact architecture is shown in Table \ref{tbl:res_convnet}).  

In addition to the four environments introduced above, here we experiment with more difficult versions of the \emph{2D Chaser} and \emph{3D KeepUp} tasks. In \emph{2D Chaser with Multi-Texture}, in each episode the background of the environment is randomly selected out of four different backgrounds (shown in Figure \ref{fig:backgrounds}). This increases the perceptual complexity of the task. 
In \emph{3D KeepUp with High Motion Penalty} the motion penalty in the reward is increased, to further reduce the overall speed and unnecessary movement of the robot.

\begin{figure}
\vspace{-2mm}
\centering
\setlength{\tabcolsep}{0.5mm}
\begin{tabular}{rrl}
\includegraphics[height=3.5cm]{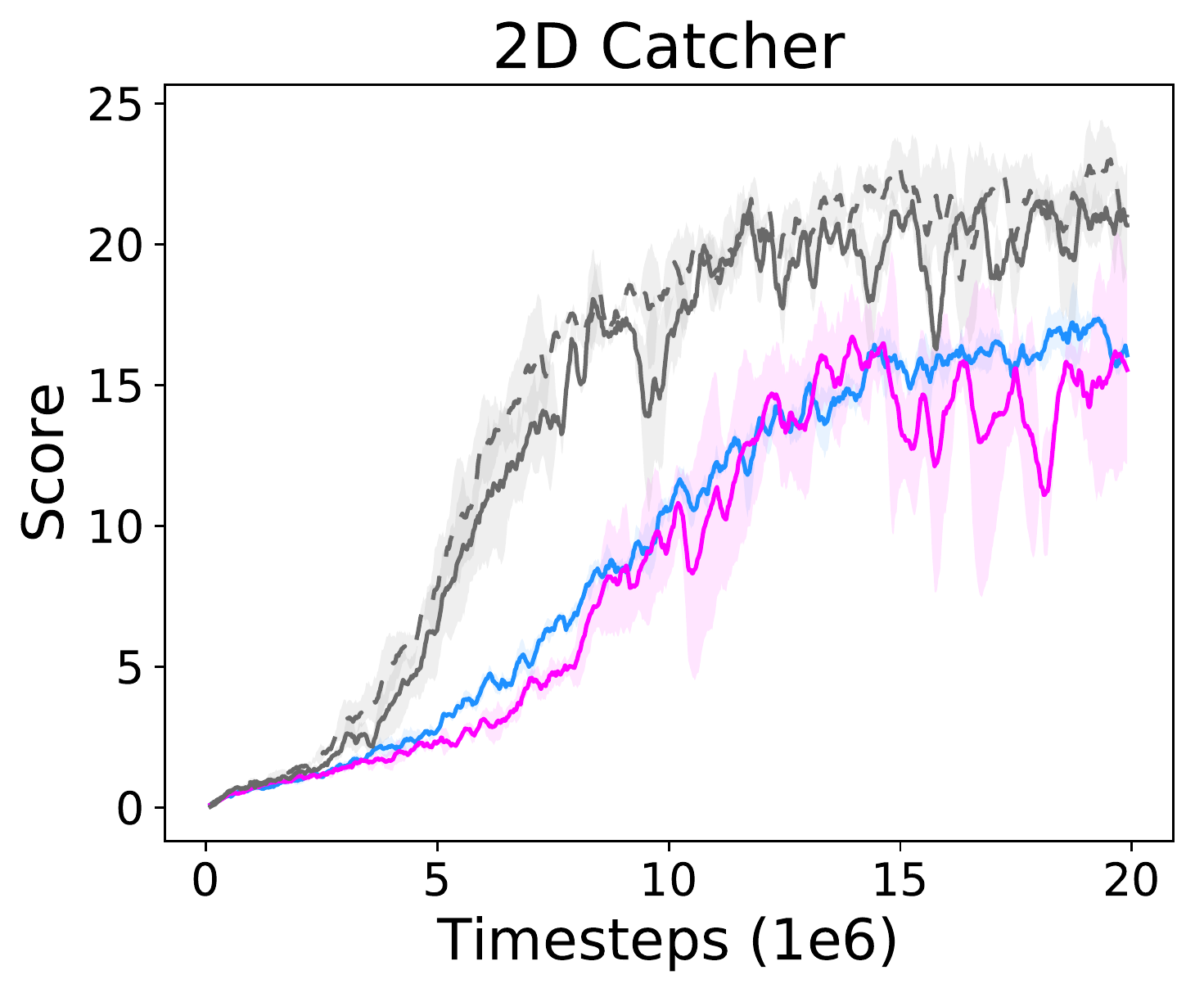} &
\includegraphics[height=3.5cm]{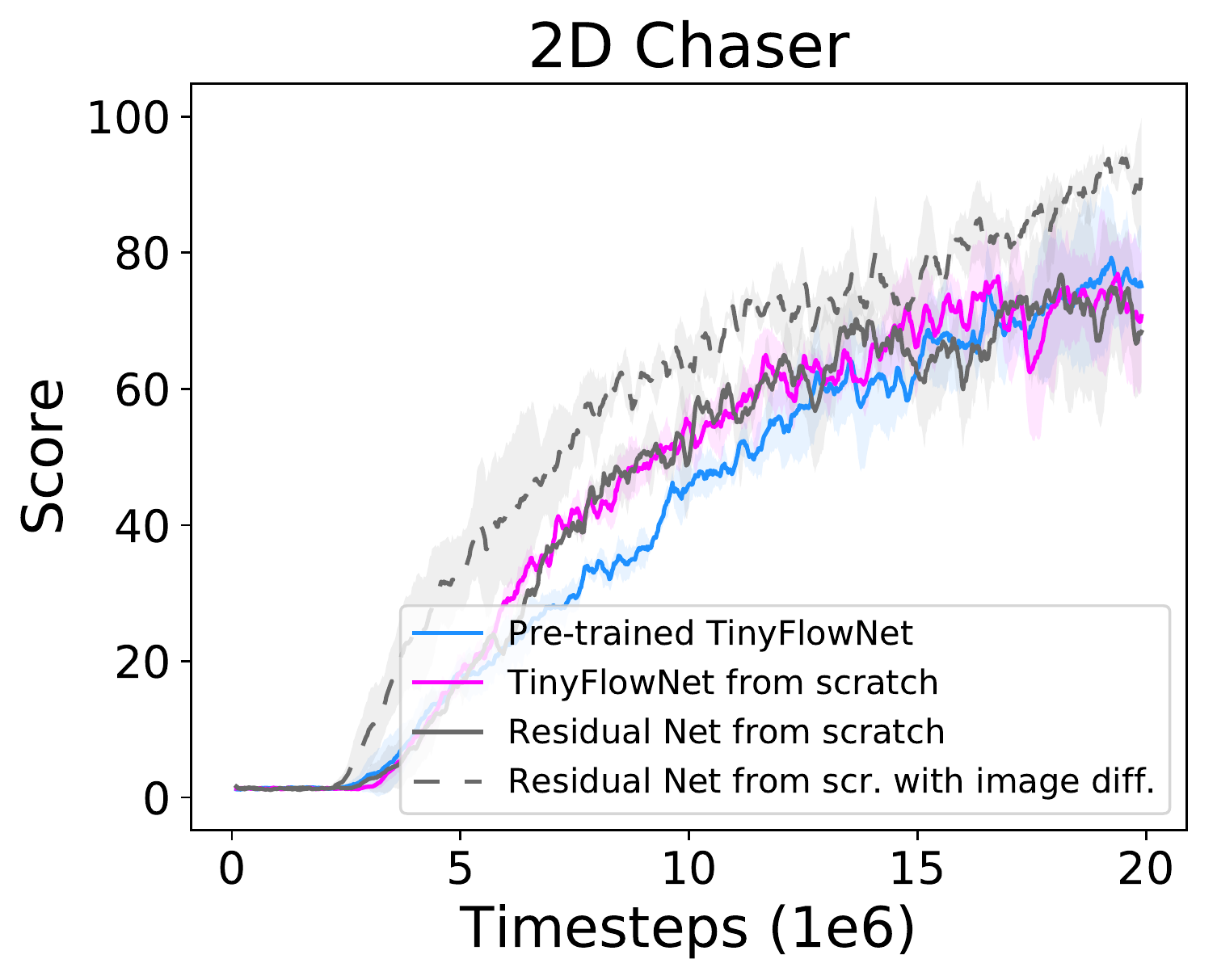} & \hspace{4mm}
\includegraphics[height=3.5cm]{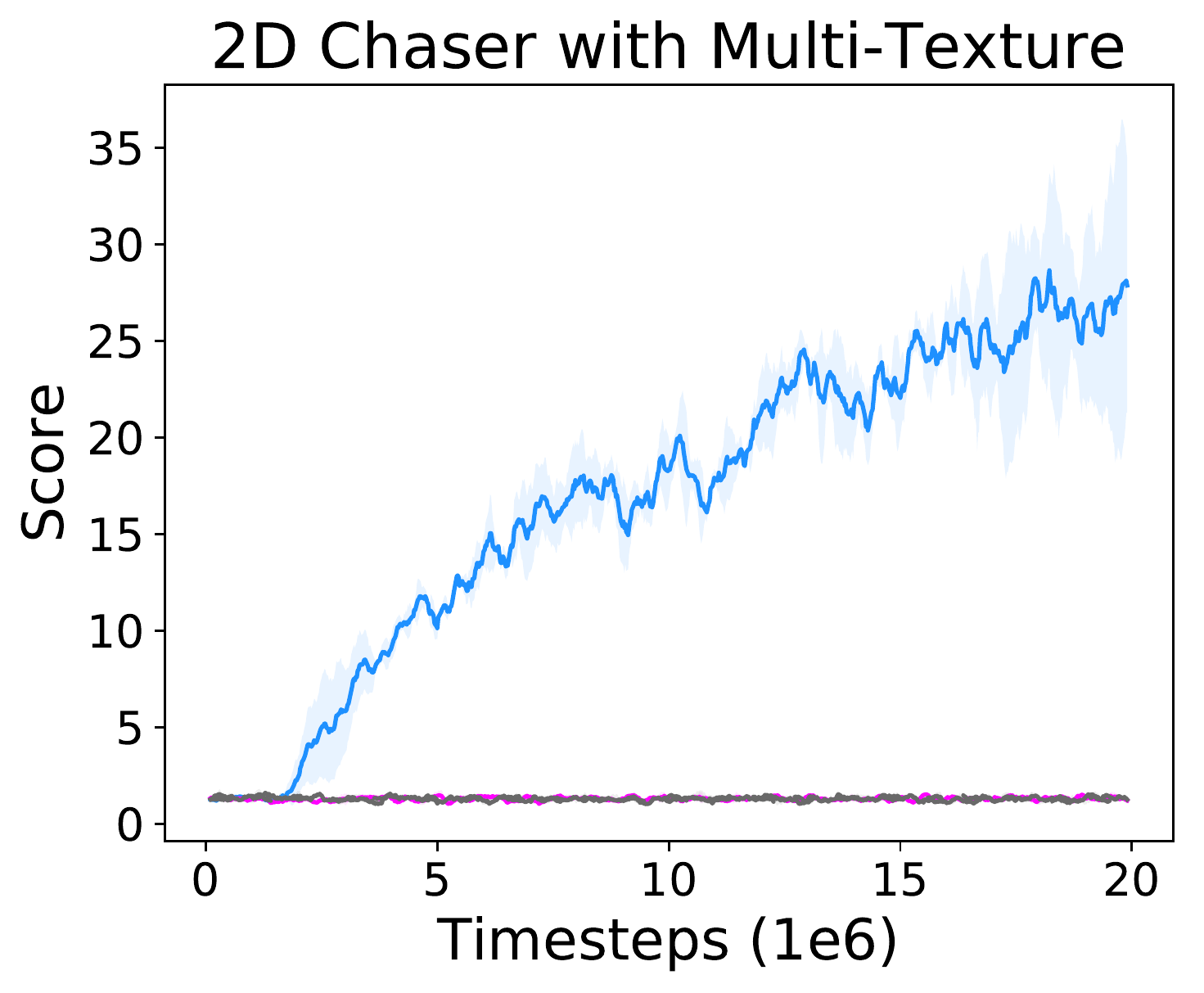} \\
\includegraphics[height=3.5cm]{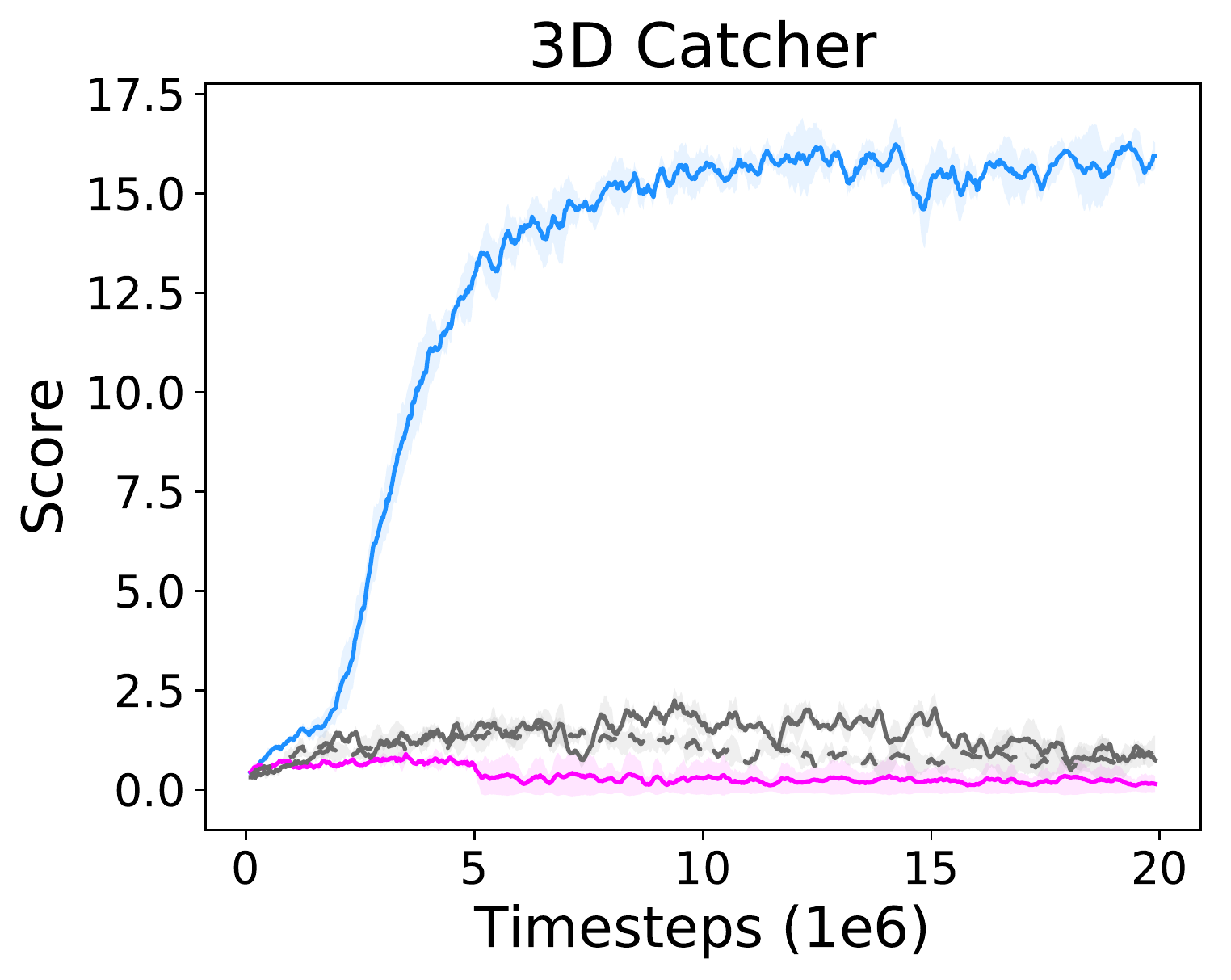} & \includegraphics[height=3.5cm]{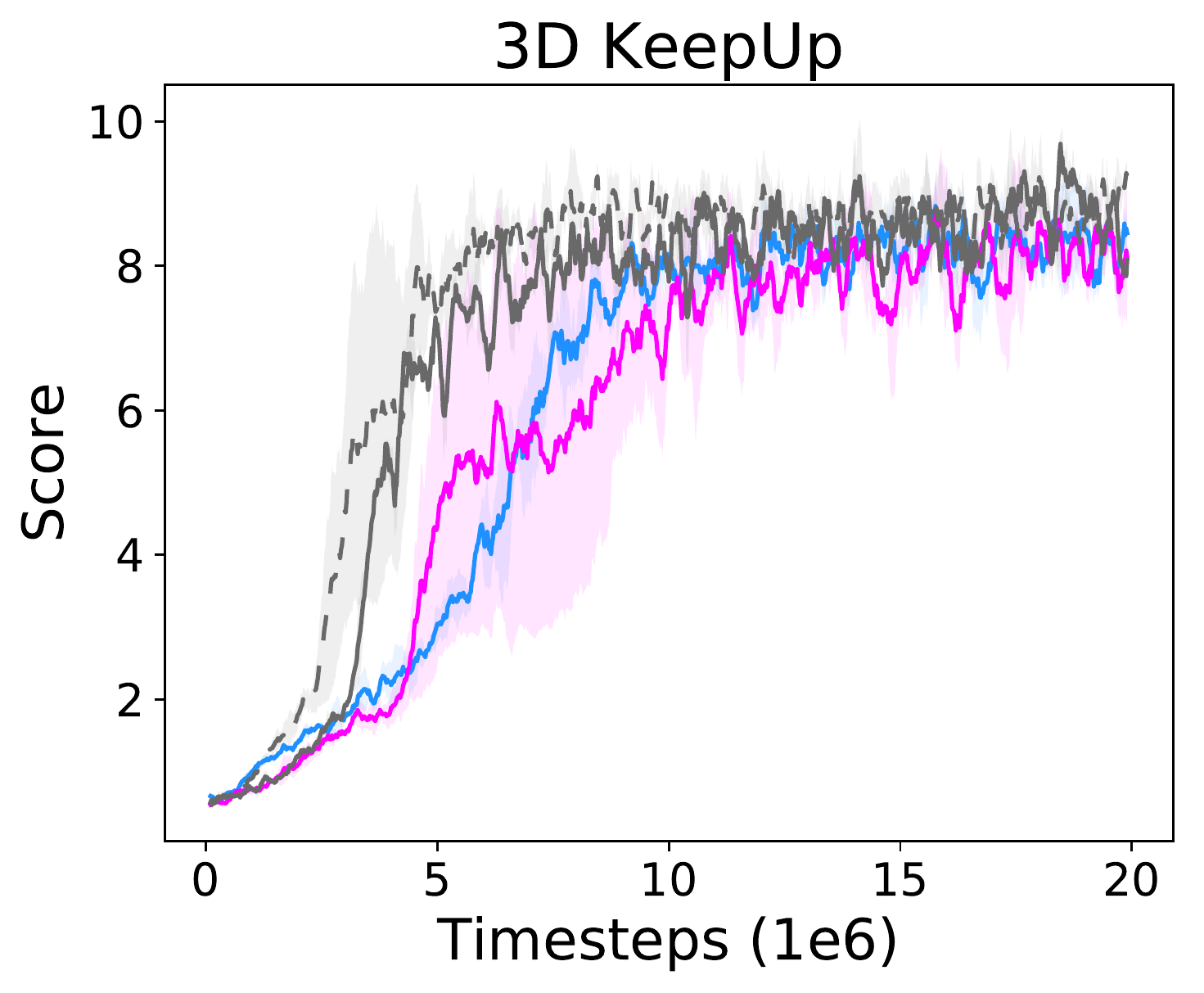} & \hspace{4.5mm}
\includegraphics[height=3.5cm]{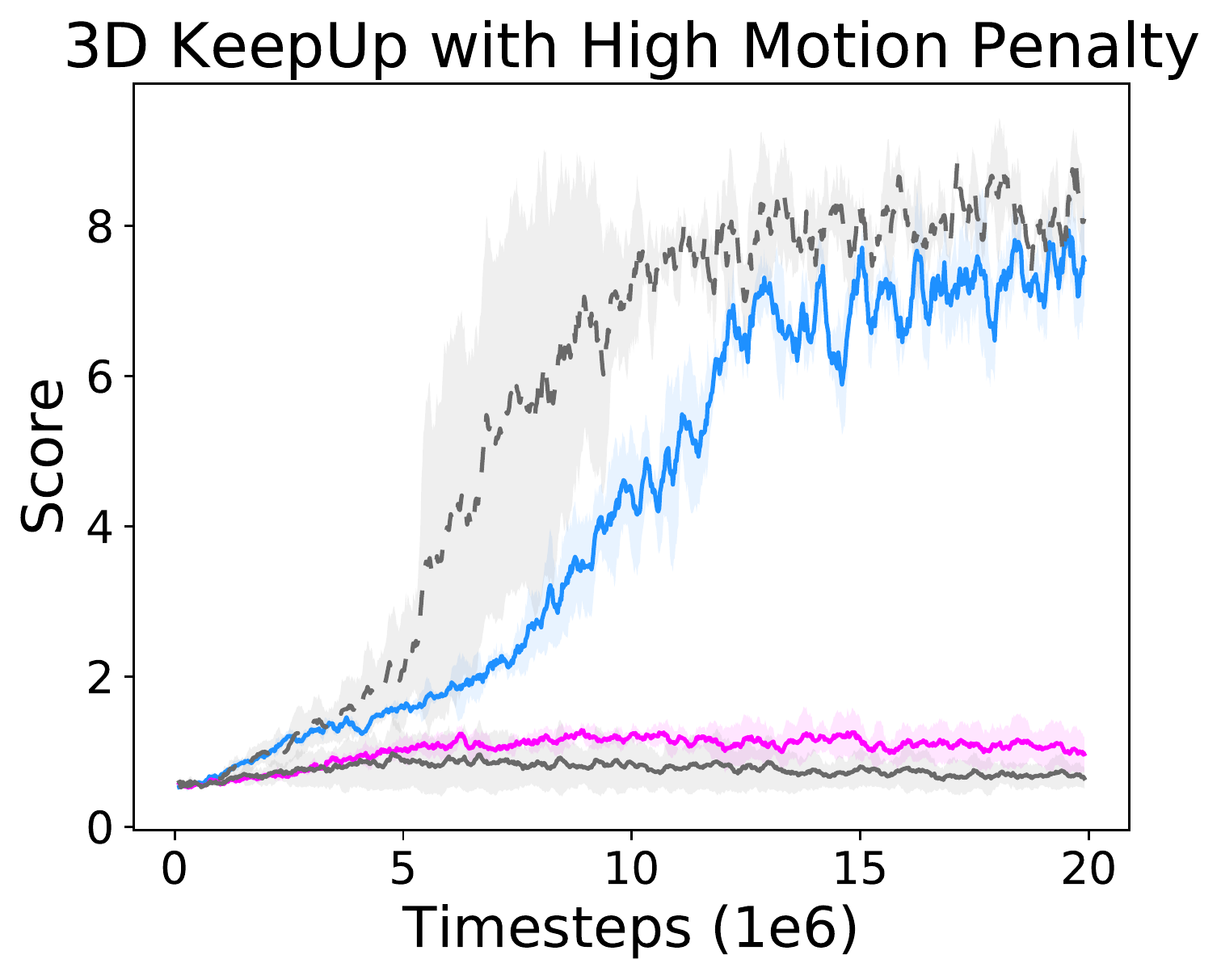} \\
\end{tabular}
        \caption{Comparison of a fixed pre-trained TinyFlowNet, a TinyFlowNet trained from scratch within the RL framework, and a deep residual RL network without the TinyFlowNet architecture.}
    	\label{fig:scratch_results}
\vspace{-3mm}
\end{figure}

The results on all six environments are shown in Figure~\ref{fig:scratch_results}.
Surprisingly, and in contrast to the architectures evaluated in the previous section, for both advanced architectures training from scratch works very well in some of the environments.
However, in the more complex tasks~-- \emph{3D Catcher}, \emph{2D Chaser with Multi-Texture}, and \emph{3D KeepUp with High Motion Penalty}~-- training from scratch with an image stack input does not yield a successful policy. In particular, in \emph{3D KeepUp with High Motion Penalty} training from scratch gets stuck in a local optimum of not moving the robot arm, while the agent with pre-trained TinyFlowNet is still able to solve the task. Providing the residual network with the image difference improves its performance on the \emph{2D Chaser} tasks and results in a successful policy on the \emph{3D KeepUp with High Motion Penalty}. This indicates that using the image difference as an additional input also improves the performance of larger architectures.

Overall, although in several cases a larger architecture can learn the necessary motion features based only on the reward signal, the use of a pre-trained optical flow estimator is still beneficial and allows for robust training on a wider range of environments.

The two advanced architectures trained from scratch reach similar scores in all environments; however, the architecture including TinyFlowNet has the advantage of being more interpretable, since it predicts an intermediate optical-flow-like two-channel representation.
We show example outputs of an automatically learned TinyFlowNet in Figure~\ref{fig:scratch_flow}.
To visualize the two-channel outputs of the network, we assign them to two color channels of an RGB image: red and blue.
Interestingly, the network learned to represent the motion of the ball and largely ignore the motion of the robotic arm.
The representation of the motion generated by the network is different from the standard optical flow representation: instead of encoding the (x,y) displacements in the two channels of the result, the network displaces the content of the two channels spatially in the direction of the motion.

\section{Conclusion}
\label{sec:conclusion}

In this work we showcased the importance of an explicit motion representation for control tasks that involve dynamic objects. 
We presented the integration of an optical flow network into a reinforcement learning setup and showed that the use of optical flow helped on tasks that involve dynamics. 
Interestingly, on several tasks, motion features were learned in an unsupervised manner just from task-specific rewards and achieved the same and sometime higher performance than the network that was trained to predict optical flow in a supervised manner. 
On two tasks, unsupervised learning was not successful and kickstarting the use of motion by supervised learning of optical flow was necessary. 

Further, we found that in all experiments providing image difference as input to the network matched or outperformed the image stack input.
RL with image difference input was not able to solve all of the tasks with dynamic objects, however, the faster computation time compared to optical flow estimation makes it a viable alternative for tasks with simple motion components. We still expect that the image difference will not outperform the image stack in environments with more complex motion, including large displacements or egomotion.
Overall our results suggest using image difference as the default input representation instead of an image stack when performing RL in dynamic environments. 

Our work opens up several opportunities for future research.
First, it would be interesting to apply similar methods to more complex environments and eventually to physical robotic systems.
We expect that pre-trained perception systems would be even more beneficial in these more complex conditions, and, moreover, the use of the abstract optical flow representation may simplify the transfer from simulation to the real world~\citep{Clavera2017,Mueller2018}.
Second, rather than pre-training optical flow using supervised learning, one could use unsupervised methods based on frame prediction~\citep{Finn2016,Yu2016}.
Third, learning of motion features just from rewards in several tasks is interesting by itself and only succeeded due to the deeper network architectures. 
How the use of suitable network architectures may generally help improve representation learning in control setups is worth further investigation.


\acknowledgments{We thank Sergio Guadarrama for suggesting the image difference approach~\cite{wang2016temporal} and Max Argus for useful discussions and for help with preparation of the manuscript. 
This project was funded in part by the BrainLinks-BrainTools Cluster of Excellence (DFG EXC 1086) and by the Intel Network on Intelligent Systems.}



\bibliography{main}

\begin{thebibliography}{35}
\providecommand{\natexlab}[1]{#1}
\providecommand{\url}[1]{\texttt{#1}}
\expandafter\ifx\csname urlstyle\endcsname\relax
  \providecommand{\doi}[1]{doi: #1}\else
  \providecommand{\doi}{doi: \begingroup \urlstyle{rm}\Url}\fi

\bibitem[Allen et~al.(1991)Allen, Yoshimi, and Timcenko]{Allen1991}
P.~K. Allen, B.~Yoshimi, and A.~Timcenko.
\newblock Real-time visual servoing.
\newblock In \emph{ICRA}, 1991.

\bibitem[Andrychowicz et~al.(2017)Andrychowicz, Wolski, Ray, Schneider, Fong,
  Welinder, McGrew, Tobin, Pieter~Abbeel, and Zaremba]{Andrychowicz2017}
M.~Andrychowicz, F.~Wolski, A.~Ray, J.~Schneider, R.~Fong, P.~Welinder,
  B.~McGrew, J.~Tobin, O.~Pieter~Abbeel, and W.~Zaremba.
\newblock Hindsight experience replay.
\newblock In \emph{NIPS}. 2017.

\bibitem[Brockman et~al.(2016)Brockman, Cheung, Pettersson, Schneider,
  Schulman, Tang, and Zaremba]{brockman2016openai}
G.~Brockman, V.~Cheung, L.~Pettersson, J.~Schneider, J.~Schulman, J.~Tang, and
  W.~Zaremba.
\newblock Openai gym.
\newblock \emph{arXiv preprint arXiv:1606.01540}, 2016.

\bibitem[Chao et~al.(2014)Chao, Gu, and Napolitano]{Chao2014}
H.~Chao, Y.~Gu, and M.~Napolitano.
\newblock A survey of optical flow techniques for robotics navigation
  applications.
\newblock \emph{Journal of Intelligent {\&} Robotic Systems}, 2014.

\bibitem[Clavera et~al.(2017)Clavera, Held, and Abbeel]{Clavera2017}
I.~Clavera, D.~Held, and P.~Abbeel.
\newblock Policy transfer via modularity and reward guiding.
\newblock In \emph{IROS}, 2017.

\bibitem[Dosovitskiy and Koltun(2017)]{DosovitskiyKoltun2017}
A.~Dosovitskiy and V.~Koltun.
\newblock Learning to act by predicting the future.
\newblock In \emph{International Conference on Learning Representations}, 2017.

\bibitem[Dosovitskiy et~al.(2015)Dosovitskiy, Fischer, Ilg, H{\"a}usser,
  Haz{\i}rba{\c{s}}, Golkov, v.d. Smagt, Cremers, and Brox]{Dosovitskiy2015}
A.~Dosovitskiy, P.~Fischer, E.~Ilg, P.~H{\"a}usser, C.~Haz{\i}rba{\c{s}},
  V.~Golkov, P.~v.d. Smagt, D.~Cremers, and T.~Brox.
\newblock Flownet: Learning optical flow with convolutional networks.
\newblock In \emph{International Conference on Computer Vision}, 2015.

\bibitem[Dwibedi et~al.(2018)Dwibedi, Tompson, Lynch, and
  Sermanet]{dwibedi2018learning}
D.~Dwibedi, J.~Tompson, C.~Lynch, and P.~Sermanet.
\newblock Learning actionable representations from visual observations.
\newblock In \emph{IEEE/RSJ International Conference on Intelligent Robots and
  Systems (IROS)}, 2018.

\bibitem[Ebert et~al.(2017)Ebert, Finn, Lee, and Levine]{Ebert2017}
F.~Ebert, C.~Finn, A.~X. Lee, and S.~Levine.
\newblock Self-supervised visual planning with temporal skip connections.
\newblock In \emph{Conference on Robot Learning}, 2017.

\bibitem[Finn and Levine(2017)]{FinnLevine2017}
C.~Finn and S.~Levine.
\newblock Deep visual foresight for planning robot motion.
\newblock In \emph{ICRA}, 2017.

\bibitem[Finn et~al.(2016)Finn, Goodfellow, and Levine]{Finn2016}
C.~Finn, I.~J. Goodfellow, and S.~Levine.
\newblock Unsupervised learning for physical interaction through video
  prediction.
\newblock In \emph{NIPS}, 2016.

\bibitem[Goel et~al.(2018)Goel, Weng, and Poupart]{Goel2018}
V.~Goel, J.~Weng, and P.~Poupart.
\newblock Unsupervised video object segmentation for deep reinforcement
  learning.
\newblock \emph{arxiv:1805.07780}, 2018.

\bibitem[Gu et~al.(2017)Gu, Holly, Lillicrap, and Levine]{Gu2017}
S.~Gu, E.~Holly, T.~Lillicrap, and S.~Levine.
\newblock Deep reinforcement learning for robotic manipulation with
  asynchronous off-policy updates.
\newblock In \emph{ICRA}, 2017.

\bibitem[He et~al.(2016)He, Zhang, Ren, and Sun]{he2016deep}
K.~He, X.~Zhang, S.~Ren, and J.~Sun.
\newblock Deep residual learning for image recognition.
\newblock In \emph{IEEE Conference on Computer Vision and Pattern Recognition
  (CVPR)}, 2016.

\bibitem[Henderson et~al.(2017)Henderson, Islam, Bachman, Pineau, Precup, and
  Meger]{henderson2017deep}
P.~Henderson, R.~Islam, P.~Bachman, J.~Pineau, D.~Precup, and D.~Meger.
\newblock Deep reinforcement learning that matters.
\newblock \emph{arXiv preprint arXiv:1709.06560}, 2017.

\bibitem[Ilg et~al.(2017)Ilg, Mayer, Saikia, Keuper, Dosovitskiy, and
  Brox]{IMSKDB17}
E.~Ilg, N.~Mayer, T.~Saikia, M.~Keuper, A.~Dosovitskiy, and T.~Brox.
\newblock Flownet 2.0: Evolution of optical flow estimation with deep networks.
\newblock In \emph{IEEE Conference on Computer Vision and Pattern Recognition
  (CVPR)}, 2017.

\bibitem[Jaderberg et~al.(2017)Jaderberg, Mnih, Czarnecki, Schaul, Leibo,
  Silver, and Kavukcuoglu]{Jaderberg2017}
M.~Jaderberg, V.~Mnih, W.~M. Czarnecki, T.~Schaul, J.~Z. Leibo, D.~Silver, and
  K.~Kavukcuoglu.
\newblock Reinforcement learning with unsupervised auxiliary tasks.
\newblock In \emph{International Conference on Learning Representations}, 2017.

\bibitem[Lillicrap et~al.(2016)Lillicrap, Hunt, Pritzel, Heess, Erez, Tassa,
  Silver, and Wierstra]{Lillicrap2016}
T.~P. Lillicrap, J.~J. Hunt, A.~Pritzel, N.~Heess, T.~Erez, Y.~Tassa,
  D.~Silver, and D.~Wierstra.
\newblock Continuous control with deep reinforcement learning.
\newblock In \emph{International Conference on Learning Representations}, 2016.

\bibitem[Luo et~al.(1988)Luo, Mullen, and Wessell]{Luo1988}
R.~C. Luo, R.~E. Mullen, and D.~E. Wessell.
\newblock An adaptive robotic tracking system using optical flow.
\newblock In \emph{ICRA}, 1988.

\bibitem[Mirowski et~al.(2017)Mirowski, Pascanu, Viola, Soyer, Ballard, Banino,
  Denil, Goroshin, Sifre, Kavukcuoglu, Kumaran, and Hadsell]{Mirowski2017}
P.~Mirowski, R.~Pascanu, F.~Viola, H.~Soyer, A.~J. Ballard, A.~Banino,
  M.~Denil, R.~Goroshin, L.~Sifre, K.~Kavukcuoglu, D.~Kumaran, and R.~Hadsell.
\newblock Learning to navigate in complex environments.
\newblock In \emph{International Conference on Learning Representations}, 2017.

\bibitem[Mnih et~al.(2015)Mnih, Kavukcuoglu, Silver, Rusu, Veness, Bellemare,
  Graves, Riedmiller, Fidjeland, Ostrovski, Petersen, Beattie, Sadik,
  et~al.]{Mnih2015}
V.~Mnih, K.~Kavukcuoglu, D.~Silver, A.~A. Rusu, J.~Veness, M.~G. Bellemare,
  A.~Graves, M.~Riedmiller, A.~K. Fidjeland, G.~Ostrovski, S.~Petersen,
  C.~Beattie, A.~Sadik, et~al.
\newblock Human-level control through deep reinforcement learning.
\newblock \emph{Nature}, 2015.

\bibitem[Mnih et~al.(2016)Mnih, Badia, Mirza, Graves, Lillicrap, Harley,
  Silver, and Kavukcuoglu]{Mnih2016}
V.~Mnih, A.~P. Badia, M.~Mirza, A.~Graves, T.~P. Lillicrap, T.~Harley,
  D.~Silver, and K.~Kavukcuoglu.
\newblock Asynchronous methods for deep reinforcement learning.
\newblock In \emph{International Conference on Machine Learning}, 2016.

\bibitem[M{\"{u}}ller et~al.(2018)M{\"{u}}ller, Dosovitskiy, Ghanem, and
  Koltun]{Mueller2018}
M.~M{\"{u}}ller, A.~Dosovitskiy, B.~Ghanem, and V.~Koltun.
\newblock Driving policy transfer via modularity and abstraction.
\newblock \emph{arxiv:1804.09364}, 2018.

\bibitem[Muratet et~al.(2004)Muratet, Doncieux, and Meyer]{Muratet2004}
L.~Muratet, S.~Doncieux, and J.-A. Meyer.
\newblock {A} biomimetic reactive navigation system using the optical flow for
  a rotary-wing {UAV} in urban environment.
\newblock In \emph{{I}nternational {S}ymposium on {R}obotics}, 2004.

\bibitem[Savva et~al.(2017)Savva, Chang, Dosovitskiy, Funkhouser, and
  Koltun]{Savva2017}
M.~Savva, A.~X. Chang, A.~Dosovitskiy, T.~Funkhouser, and V.~Koltun.
\newblock {MINOS}: Multimodal indoor simulator for navigation in complex
  environments.
\newblock \emph{arXiv:1712.03931}, 2017.

\bibitem[Schulman et~al.(2017)Schulman, Wolski, Dhariwal, Radford, and
  Klimov]{schulman2017proximal}
J.~Schulman, F.~Wolski, P.~Dhariwal, A.~Radford, and O.~Klimov.
\newblock Proximal policy optimization algorithms.
\newblock \emph{arXiv preprint arXiv:1707.06347}, 2017.

\bibitem[Souhila and Karim(2007)]{SouhilaKarim2007}
K.~Souhila and A.~Karim.
\newblock Optical flow based robot obstacle avoidance.
\newblock \emph{International Journal of Advanced Robotic Systems}, 2007.

\bibitem[Su and Shen(2016)]{SuShaojie2016}
K.~Su and S.~Shen.
\newblock Catching a flying ball with a vision-based quadrotor.
\newblock In \emph{ISER}, 2016.

\bibitem[Tassa et~al.(2018)Tassa, Doron, Muldal, Erez, Li, Casas, Budden,
  Abdolmaleki, Merel, Lefrancq, et~al.]{tassa2018deepmind}
Y.~Tassa, Y.~Doron, A.~Muldal, T.~Erez, Y.~Li, D.~d.~L. Casas, D.~Budden,
  A.~Abdolmaleki, J.~Merel, A.~Lefrancq, et~al.
\newblock Deepmind control suite.
\newblock \emph{arXiv preprint arXiv:1801.00690}, 2018.

\bibitem[Tobin et~al.(2017)Tobin, Fong, Ray, Schneider, Zaremba, and
  Abbeel]{Tobin2017}
J.~Tobin, R.~Fong, A.~Ray, J.~Schneider, W.~Zaremba, and P.~Abbeel.
\newblock Domain randomization for transferring deep neural networks from
  simulation to the real world.
\newblock In \emph{IROS}, 2017.

\bibitem[Todorov et~al.(2012)Todorov, Erez, and Tassa]{todorov2012mujoco}
E.~Todorov, T.~Erez, and Y.~Tassa.
\newblock Mujoco: A physics engine for model-based control.
\newblock In \emph{Intelligent Robots and Systems (IROS), 2012 IEEE/RSJ
  International Conference on}, pages 5026--5033. IEEE, 2012.

\bibitem[Vardy and Moller(2005)]{VardyMoller2005}
A.~Vardy and R.~Moller.
\newblock Biologically plausible visual homing methods based on optical flow
  techniques.
\newblock \emph{Connection Science}, 17:\penalty0 47--89, 2005.

\bibitem[Wang et~al.(2016)Wang, Xiong, Wang, Qiao, Lin, Tang, and
  Van~Gool]{wang2016temporal}
L.~Wang, Y.~Xiong, Z.~Wang, Y.~Qiao, D.~Lin, X.~Tang, and L.~Van~Gool.
\newblock Temporal segment networks: Towards good practices for deep action
  recognition.
\newblock In \emph{European Conference on Computer Vision}. Springer, 2016.

\bibitem[Xu et~al.(2017)Xu, Gao, Yu, and Darrell]{Xu2017}
H.~Xu, Y.~Gao, F.~Yu, and T.~Darrell.
\newblock End-to-end learning of driving models from large-scale video
  datasets.
\newblock In \emph{Conference on Computer Vision and Pattern Recognition},
  2017.

\bibitem[Yu et~al.(2016)Yu, Harley, and Derpanis]{Yu2016}
J.~J. Yu, A.~W. Harley, and K.~G. Derpanis.
\newblock Back to basics: Unsupervised learning of optical flow via brightness
  constancy and motion smoothness.
\newblock In \emph{{ECCV} Workshops}, 2016.

\end{thebibliography}


\section*{Supplementary Material}

\renewcommand{\thesection}{S\arabic{section}}
\setcounter{section}{0}
\renewcommand{\thefigure}{S\arabic{figure}}
\setcounter{figure}{0}
\renewcommand{\thetable}{S\arabic{table}}
\setcounter{table}{0}

\begin{figure}[h]
        \centering
            \includegraphics[width=0.435\textwidth]{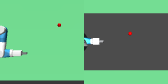}\\
            \vspace{3mm}
            \includegraphics[width=0.435\textwidth]{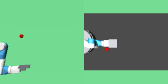}
            \caption{Side and top view of the \emph{3D Catcher} (top) and \emph{3D KeepUp} (bottom) tasks which are provided to the RL agent.} 
        \label{fig:robot_view}
\end{figure}

\begin{figure}[h]
    \centering
    { \small
    \def\arraystretch{0.5}
	\setlength{\tabcolsep}{0.3mm}
	\begin{tabular}{cc}
    \includegraphics[width=0.215\linewidth]{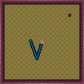} &
    \includegraphics[width=0.215\linewidth]{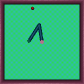}\\
    \includegraphics[width=0.215\linewidth]{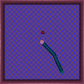} &
    \includegraphics[width=0.215\linewidth]{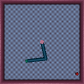}\\
    \end{tabular}
    }
    \caption{Four backgrounds of the \emph{2D Multi-Texture Chaser} task.}
    \label{fig:backgrounds}
\end{figure}

\newpage

\subsection*{Network architectures}
\begin{table}[h]
    \vspace{-1mm}
	\centering
	\ra{1.1}
	\resizebox{\linewidth}{!}{
		\small
		\begin{tabular}{ll|ccc|l}
			\toprule
			\multicolumn{1}{l}{Network part} & \multicolumn{1}{c}{Input} & \multicolumn{1}{|c}{Channels} & \multicolumn{1}{c}{Kernel} & \multicolumn{1}{c|}{Stride} & {Layer type}\\
			\midrule
			\multirow{4}{*}{Perception} &  Pixel input &  $32$ & $8 \times 8$ & $4$ & \multirow{3}{*}{Convolutions}\\
			& Previous layer &  $64$ & $4 \times 4$ & $2$ &\\
            & Previous layer &  $64$ & $3 \times 3$ & $1$ &\\
            & Previous layer & & & & Flatting\\
			\midrule
		    \multirow{3}{*}{Middle part} & Vector input &  $64$  &    & & Fully connected\\
		    & Perception output $+$ Previous layer  &    &    & & Concatenation\\
		    & Previous layer &  $64$  &    & & Fully connected\\
			\midrule
			Policy & Middle part output  &  \#actions  & & & Fully connected\\
			\midrule
			Baseline &  Middle part output    & $1$   & & & Fully connected\\
			\bottomrule
		\end{tabular}
	}
	\caption{Reinforcement Learning network architecture. Each convolution uses no padding.}
	\label{tbl:convnet}
\end{table}

\begin{table}[h]
    \vspace{-1mm}
	\centering
	\ra{1.1}
	\resizebox{\linewidth}{!}{
		\small
		\begin{tabular}{lc|cccc|l}
			\toprule
			\multicolumn{1}{c}{Input} & \multicolumn{1}{c}{Output} & \multicolumn{1}{|c}{Channels} & \multicolumn{1}{c}{Kernel} & \multicolumn{1}{c|}{Stride} & \multicolumn{1}{c}{Padding} & {Layer type}\\
			\midrule
			Pixel input & &  $64$ & $3 \times 3$ & $1$ & - & \multirow{2}{*}{Convolutions}\\
			Previous layer & skip\_1 &  $128$ & $3 \times 3$ & $2$ & - &\\
			\midrule
			Previous layer & &  $128$ & $3 \times 3$ & $1$ & 0-padding &\multirow{2}{*}{Convolutions}\\
			Previous layer & &  $128$ & $3 \times 3$ & $1$ & 0-padding &\\
            Previous layer, skip\_1 & & & & & & Summation\\
            \midrule
            Previous layer & skip\_2 &  $128$ & $3 \times 3$ & $2$ & - & Convolution\\
            \midrule
            Previous layer & &  $128$ & $3 \times 3$ & $1$ & 0-padding &\multirow{2}{*}{Convolutions}\\
            Previous layer & &  $128$ & $3 \times 3$ & $1$ & 0-padding &\\
            Previous layer, skip\_2 & & & & & & Summation\\
            \midrule
            Previous layer & &  $128$ & $3 \times 3$ & $2$ & - & Convolution\\
            Previous layer & Perception output & $110$ & & & & Fully connected \\
			\bottomrule
		\end{tabular}
	}
	\caption{Deep Perception architecture with residual connections.}
	\label{tbl:res_convnet}
\end{table}

\begin{table}[h]
    \vspace{-1mm}
	\centering
	\ra{1.1}
	\resizebox{\linewidth}{!}{
		\small
		\begin{tabular}{lc|ccc|l}
			\toprule
			\multicolumn{1}{c}{Input} & \multicolumn{1}{c}{Output} & \multicolumn{1}{|c}{Channels} & \multicolumn{1}{c}{Kernel} & \multicolumn{1}{c|}{Stride} & {Layer type}\\
			\midrule
			Pixel input & skip\_$1.0$ &  $64$ & $3 \times 3$ & $1$ & \multirow{5}{*}{Convolutions}\\
			Previous layer & &  $64$ & $3 \times 3$ & $2$ &\\
            Previous layer & skip\_$0.5$ &  $128$ & $3 \times 3$ & $1$ &\\
            Previous layer & &  $128$ & $3 \times 3$ & $2$ &\\
            Previous layer & &  $128$ & $3 \times 3$ & $1$ &\\
			Previous layer & &  $32$ & $4 \times 4$ & $2$ & Upconvolution\\
			Previous layer, skip\_$0.5$  & tmp & & & & Concatenation\\
			\midrule
			tmp & half\_resolution\_flow &  $2$ & $3 \times 3$ & $1$ & Convolution\\
			Previous layer & upsampled\_flow &  $2$ & nearest neighbor & & Upsample \\
			\midrule
			tmp & &  $16$ & $4 \times 4$ & $2$ & Upconvolution\\
			Previous layer, skip\_$1.0$ & & & & & Concatenation\\
			Previous layer & flow &  $2$ & $3 \times 3$ & $1$ & Convolution\\
			\bottomrule
		\end{tabular}
	}
	\caption{TinyFlowNet architecture. Each convolution and upconvolution uses zero padding.}
	\label{tbl:tinyflownet}
\end{table}

\newpage

\subsection*{Training TinyFlowNet details}

To train TinyFlowNet for a task, first a dataset consisting of 20000 images was created. Each image was rendered in both high (512x512) and low (84x84) resolution. We used a random policy for the standard control tasks and a stationary policy for the tasks with dynamic objects. For the 2D environment datasets the target velocities were uniformly sampled between 0.4 and 1.0. This allowed performing the 2D Catcher with varying target speed experiments and improved the overall flow prediction quality.

After the dataset was generated the flow between each two successive states was predicted using FlowNet2.0\cite{IMSKDB17} on the high resolution images. The flow predictions of FlowNet2.0 were downsampled to the low resolution of 84x84 and used as targets to train the TinyFlowNet. For the 3D environments the flow for the two views were predicted separately using FlowNet2.0. Thereafter the TinyFlowNet was trained to predict the flow from both views at the same time.

The TinyFlowNet was trained for 600000 steps using a batch size of 8 and the Adam optimizer (with $\beta_1 = 0.9$, $\beta_2 = 0.999$, and $\eps = 1 \times 10^{-8}$). The initial learning rate was set to $1 \times 10^{-4}$ and was reduced by half every 100000 steps. The TinyFlowNet predicts the flow first at half resolution (42x42) and then at full resolution (shown in Table \ref{tbl:tinyflownet}). The half resolution was upsampled with nearest neighbor upsampling. Both the full-resolution flow predictions ($\text{F}_x$ and $\text{F}_y$ for the horizontal and vertical flow predictions) and the upsampled flow predictions (${\text{upF}_x}$ and ${\text{upF}_y}$) are used in the loss function:

\begin{align*}
\text{FlowLoss}_{2D} = 100\cdot \frac{1}{8*84*84} \cdot
\sum_{i = 1}^{8*84*84} (
\sqrt{({\text{F}_x}_i - {{\text{F}_x}_\text{target}}_i)^2 + 
({\text{F}_y}_i - {{\text{F}_y}_\text{target}}_i)^2}  + \\
0.5 * \sqrt{({\text{upF}_x}_i - {{\text{F}_x}_\text{target}}_i)^2 + 
({\text{upF}_y}_i - {{\text{F}_y}_\text{target}}_i)^2} )
\end{align*}

The sum over $i$ in the loss iterates over each pixel of each flow prediction in the batch. For the 3D environments this sum included both the side and the top view:

\begin{align*}
\text{FlowLoss}_{3D} = 50\cdot \frac{1}{8*84*84} \cdot
\sum_{i = 1}^{8*84*84} (
\sqrt{({\text{F}_x}_i - {{\text{F}_x}_\text{target}}_i)^2 + 
({\text{F}_y}_i - {{\text{F}_y}_\text{target}}_i)^2}  + \\
0.5 * \sqrt{({\text{upF}_x}_i - {{\text{F}_x}_\text{target}}_i)^2 + 
({\text{upF}_y}_i - {{\text{F}_y}_\text{target}}_i)^2} + \\
\sqrt{({{\text{F}_\text{top}}_x}_i - {{{\text{F}_\text{top}}_x}_\text{target}}_i)^2 + 
({{\text{F}_\text{top}}_y}_i - {{{\text{F}_\text{top}}_y}_\text{target}}_i)^2}  + \\
0.5 * \sqrt{({{\text{upF}_\text{top}}_x}_i - {{{\text{F}_\text{top}}_x}_\text{target}}_i)^2 + 
({{\text{upF}_\text{top}}_y}_i - {{{\text{F}_\text{top}}_y}_\text{target}}_i)^2})
\end{align*}

The inference after the training only uses the full-resolution flow prediction.

In every environment the flow is predicted between the current and the previous frame. There are two exceptions. Because of the low simulation time-step of the Walker2D environment the agent movement between two frames is very small. Therefore for the Walker2D we estimated the flow between the current frame and the frame four steps in the past instead of the flow of successive states. The second exception is the 2D Catcher environment with varying target speed in the lowest speed setting of $0.25$. There we estimated the flow between the current frame and the frame two steps in the past.

\begin{figure}[h!]
    \centering
    { 
    \def\arraystretch{0.5}
	\setlength{\tabcolsep}{0.3mm}
	\begin{tabular}{ccc}
    Image pair & Prediction & Difference img \vspace{1mm}\\
    \includegraphics[width=0.24\linewidth]{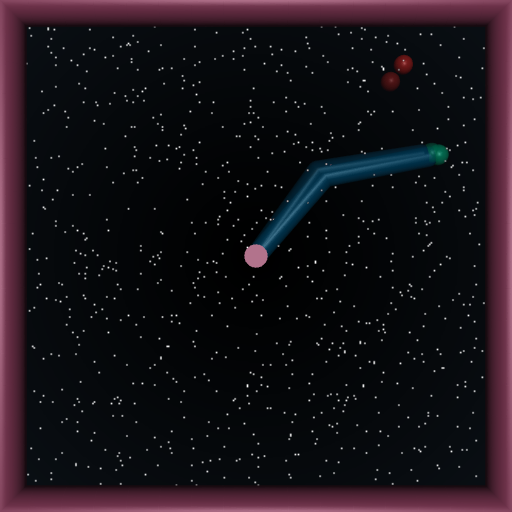} &
    \includegraphics[width=0.24\linewidth]{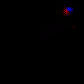} &
    \includegraphics[width=0.24\linewidth]{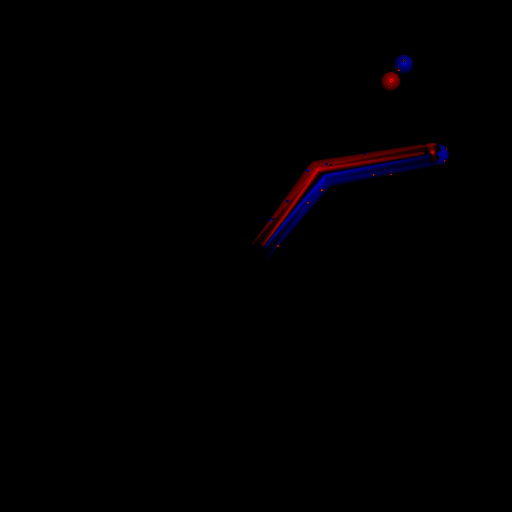} \\
    \includegraphics[width=0.24\linewidth]{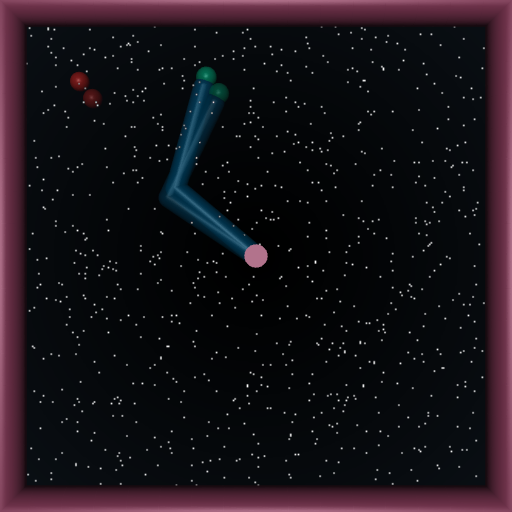} &
    \includegraphics[width=0.24\linewidth]{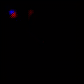} &
    \includegraphics[width=0.24\linewidth]{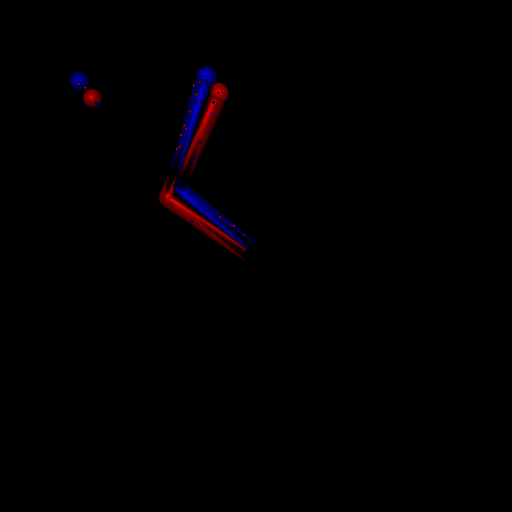} \\
    \end{tabular}
    }
    \caption{Example outputs of a TinyFlowNet trained from scratch with RL on the \emph{2D Chaser} task. Note how the moving object is clearly detected and the predicted values change depending on the motion of the object. 
    The two images on the right show the difference of the two frames. 
    To make the difference-images most similar to the prediction of the network, we subtract the grayscale versions of the two frames and assign positive values of the result to the red channel and negative values to the blue channel.
    In contrast to naive image difference, the network mostly ignores the motion of the arm.}
    \label{fig:scratch_flow}
\end{figure}

\begin{figure}[h!]
	\centering
	\setlength{\tabcolsep}{1mm}
	\begin{tabular}{rr}
	\includegraphics[height=4.2cm]{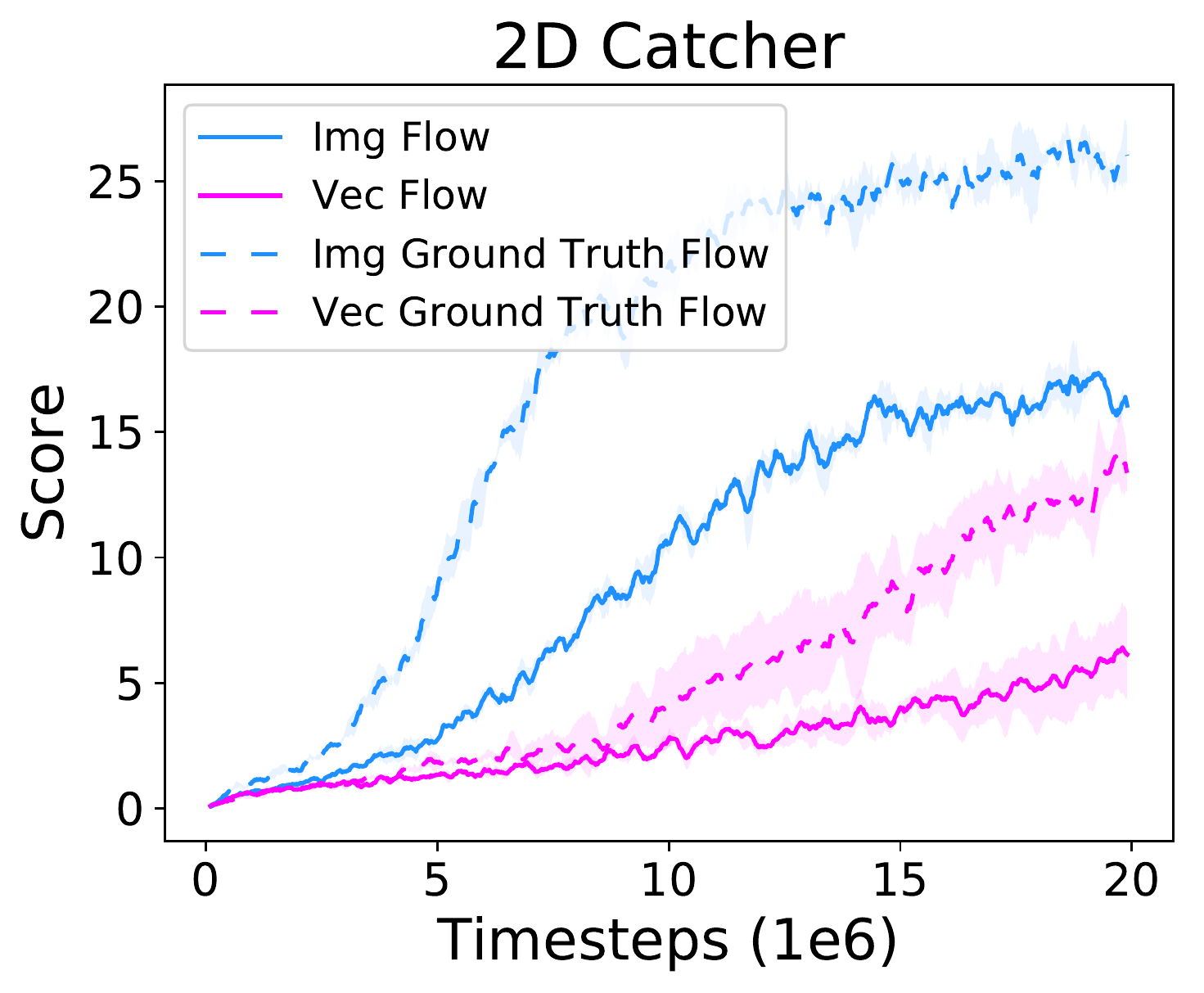} &
    \includegraphics[height=4.2cm]{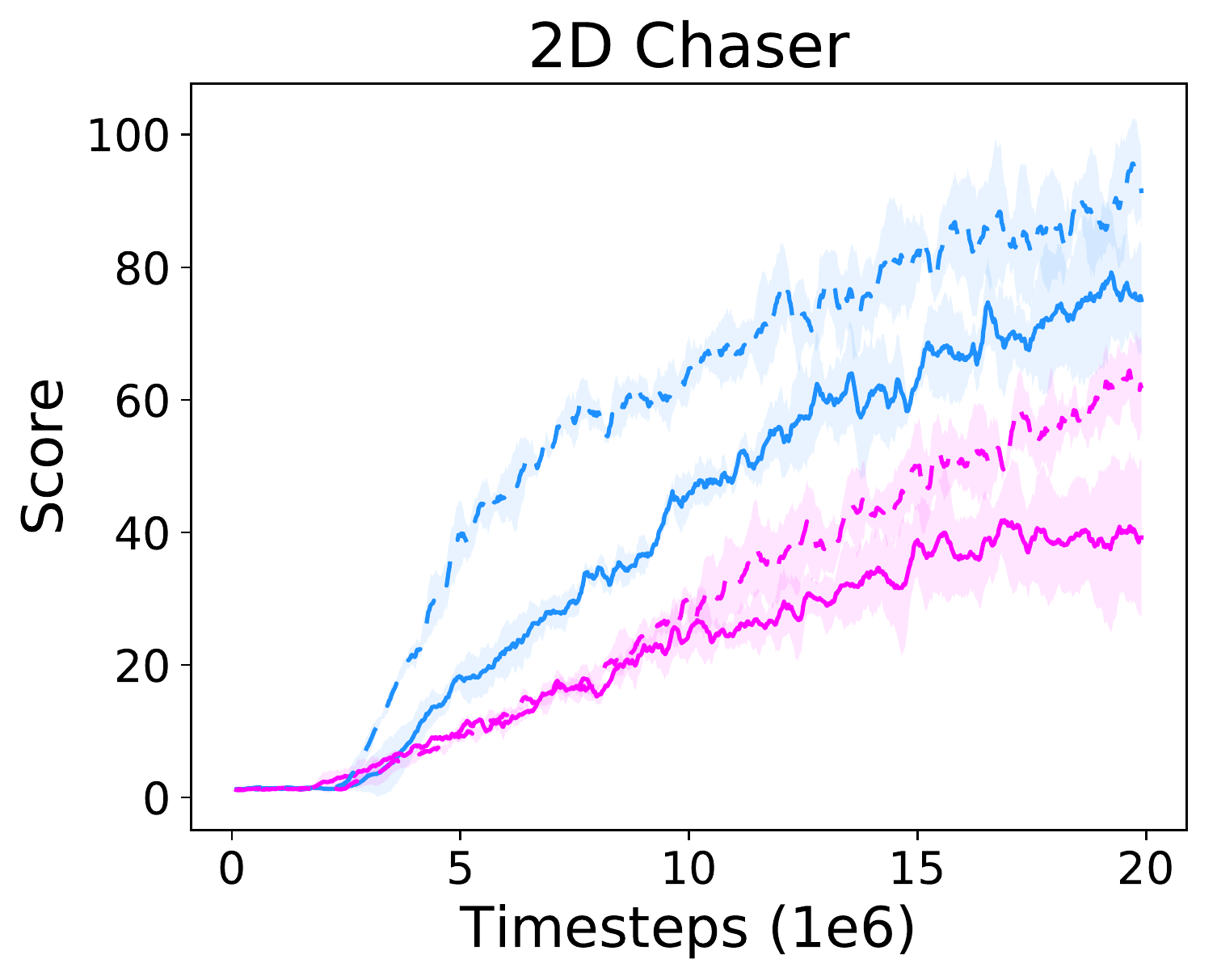} \\
    \includegraphics[height=4.2cm]{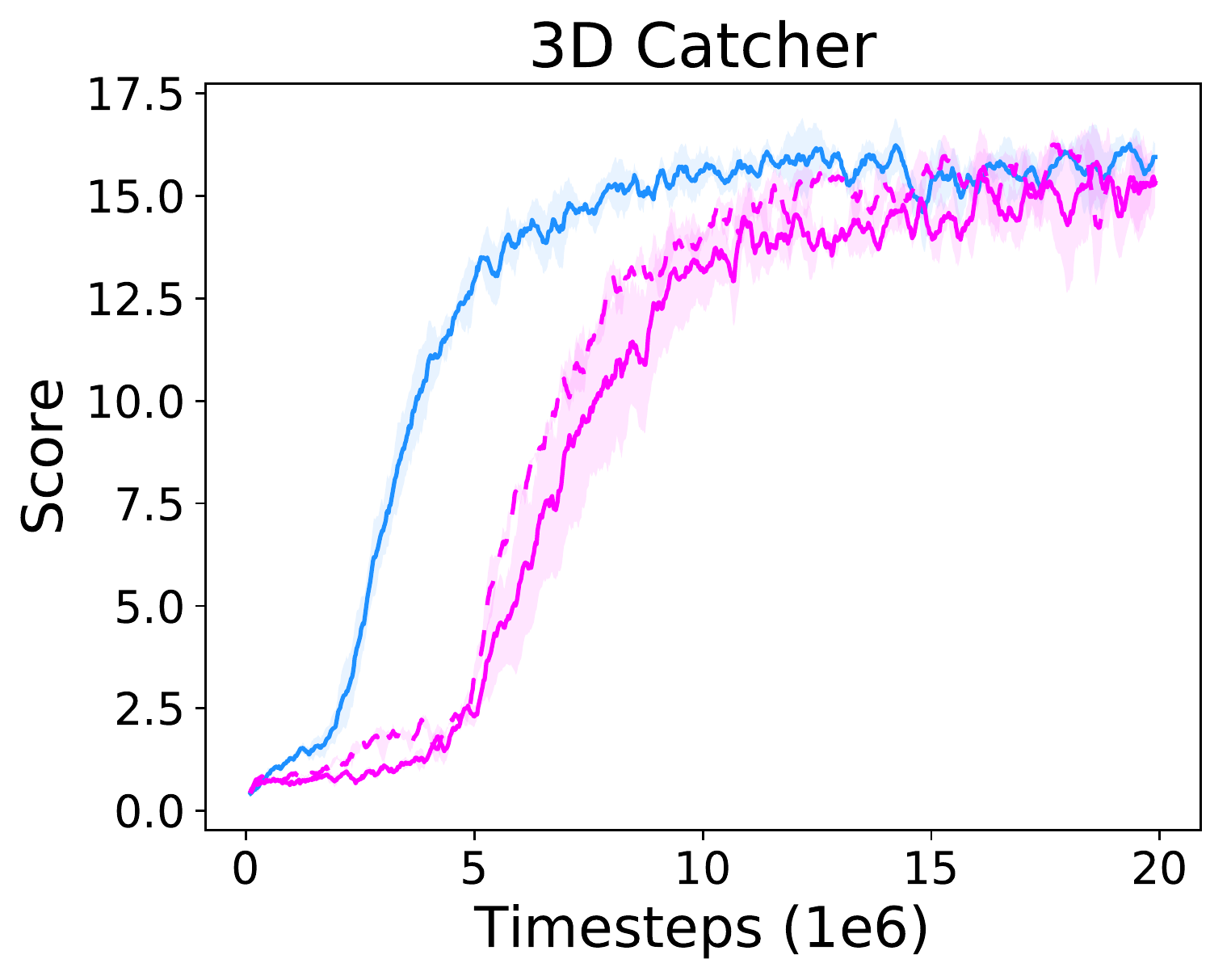} &
    \includegraphics[height=4.2cm]{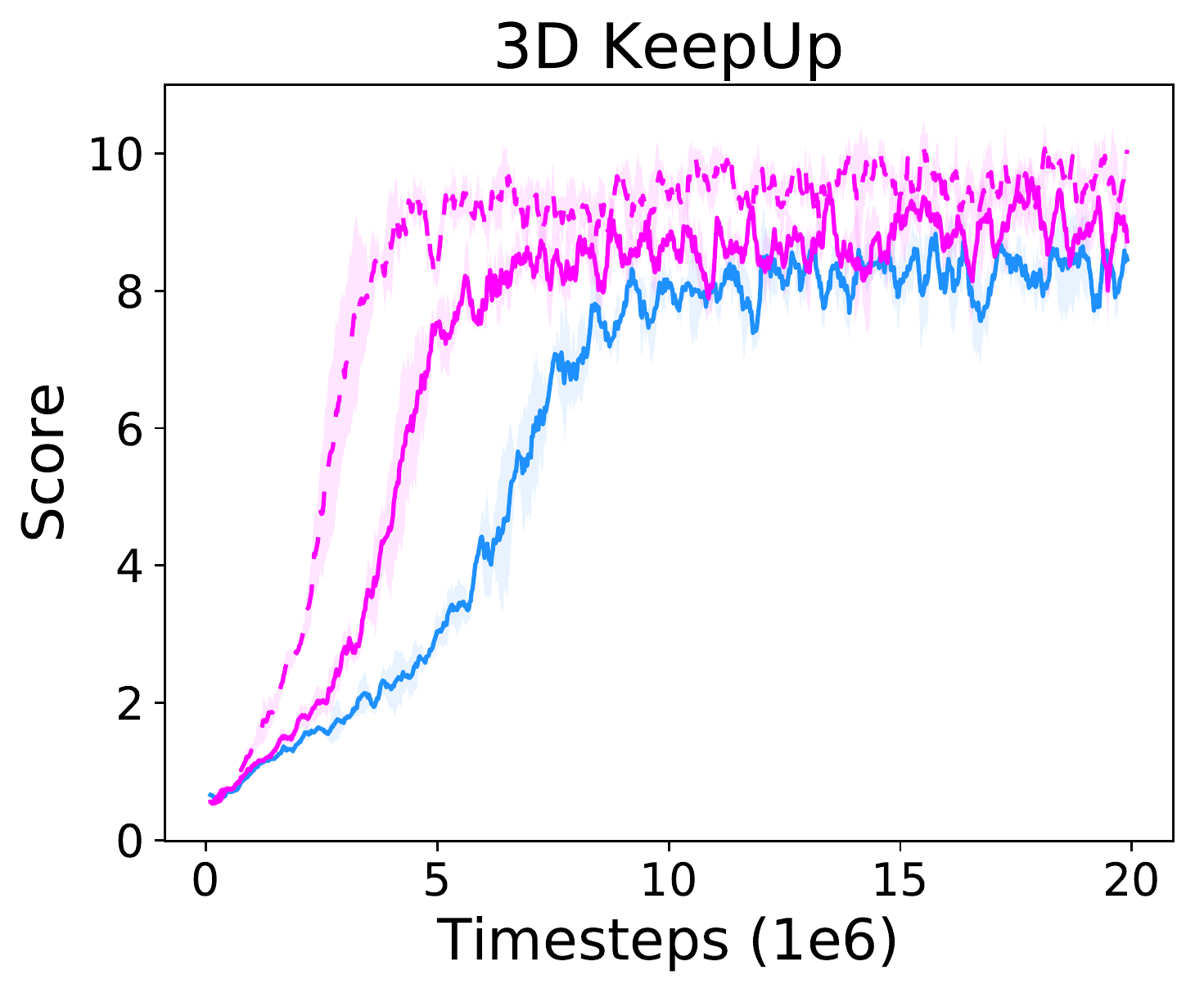} \\
    \end{tabular}
	\caption{Comparison between different motion representations. \emph{Image Flow} uses the optical flow as an additional pixel input. \emph{Vector Flow} extracts the velocity vector of the target from the optical flow by taking the average of the 6 largest flow values in each dimension. The velocity vector is then used as an additional input to the agent. The \emph{Vector Flow} approach is not easily applicable to tasks with more complex structure of motion, such as standard MuJoCo control tasks. The dashed lines show the performance of an RL agent that has been provided with ground truth optical flow instead of the TinyFlowNet prediction. We calculated the pixel optical flow ground truth only for the 2d environments. The ground truth velocity vectors are taken directly form the simulation.}
	\label{fig:flow_vs_motion_vector_results}
\end{figure}

\end{document}